\RequirePackage{fix-cm}
\documentclass[twocolumn]{svjour3}          
\smartqed  
\usepackage[colorlinks, bookmarks=false]{hyperref}
\usepackage{graphicx}
\usepackage{multirow}
\usepackage{color,xcolor}
\usepackage{amssymb}
\usepackage{amsmath}
\usepackage{wrapfig} 

\newcommand{\zjq}[1]{\textcolor[rgb]{0.0,0.0,0.0}{#1}}

\journalname{CGI2021} 
\begin{document}

\title{Multi-domain Collaborative Feature Representation for Robust Visual Object Tracking}
\subtitle{}
\author{Anonymous submission}
\institute{Anonymous submission}

\author{Jiqing Zhang$^{1,\dag}$ \and Kai Zhao$^{2,\dag}$ \and Bo Dong$^3$ \and Yingkai Fu$^{1}$ \and Yuxin Wang$^1$  \and Xin Yang$^{1,*}$ \and Baocai Yin$^1$}
\institute{Jiqing Zhang, Yingkai Fu, Yuxin Wang, Xin Yang and Baocai Yin \at
	Email: jqz@mail.dlut.edu.cn, yingkaifu@mail.dlut.edu.cn, wyx@dlut.edu.cn, xinyang@dlut.edu.cn and ybc@dlut.edu.cn \and
	Kai Zhao \at Email: kzhao@aiit.org.cn \and Bo Dong \at Email: bo.dong@sri.com \and 
	$^1$  Dalian University of Technology \and $^2$  Advanced Institute of Information Technology, Peking University \and $^3$  SRI International \at
	$^\dag$ indicates equal contribution. \at
	$^*$ Xin Yang is the Corresponding Author.}

\date{ }

\maketitle

\begin{abstract}
Jointly exploiting multiple different yet complementary domain information has been proven to be an effective way to perform robust object tracking. This paper focuses on effectively
representing and utilizing complementary features from the frame domain and event domain for boosting object tracking performance in challenge scenarios.
Specifically, we propose Common Features Extractor (CFE) to learn potential common representations from the RGB domain and event domain. For learning the unique features of the two domains, we utilize a Unique Extractor for Event (UEE)  based on Spiking Neural Networks to extract edge cues in the event domain which may be missed in RGB in some challenging conditions, and a Unique Extractor for RGB (UER)  based on Deep Convolutional Neural Networks to extract texture and semantic information in RGB domain. 
Extensive experiments on standard RGB benchmark and real event tracking dataset demonstrate the effectiveness of the proposed approach. We show our approach outperforms all compared state-of-the-art tracking algorithms and verify event-based data is a powerful cue for tracking in challenging scenes.
\end{abstract}

\section{\zjq{Introduction}}

\def\wdenoising{0.32\linewidth}
\def\hdenoising{0.8in}
\def\wcolor{0.9\linewidth}
\def\hcolor{0.26in}
\begin{figure}[t]
	\setlength{\tabcolsep}{2.4pt}
	\centering
	\begin{tabular}{ccc}

		\includegraphics[width=\wdenoising, height=\hdenoising]{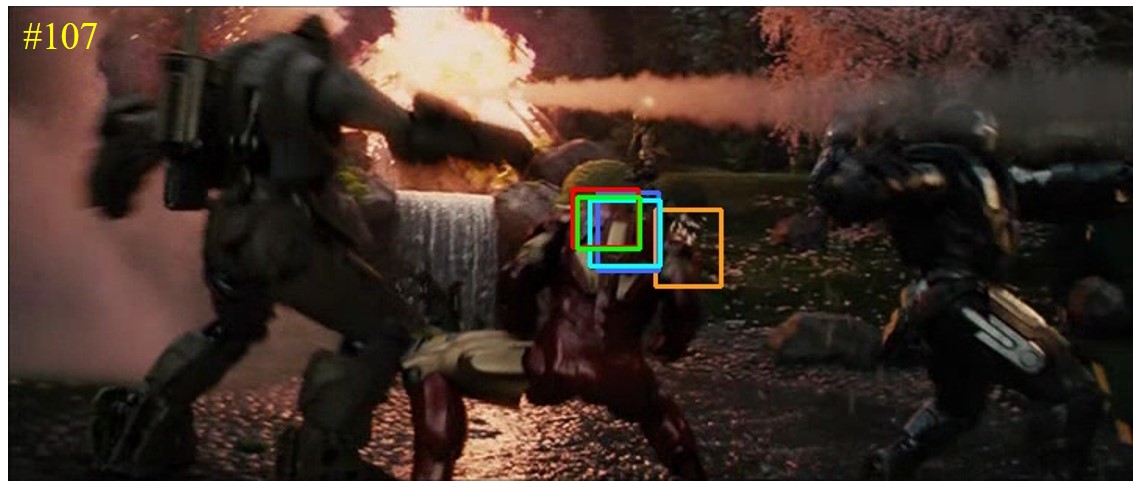}&
		\includegraphics[width=\wdenoising, height=\hdenoising]{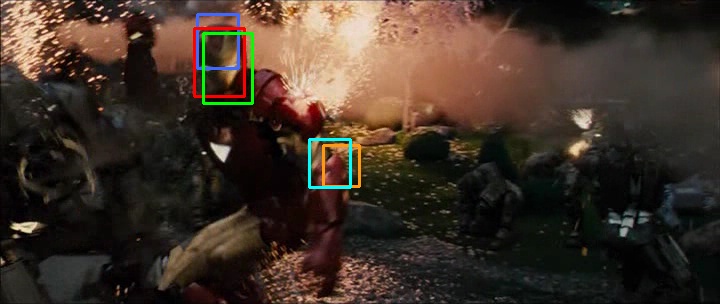}&
		\includegraphics[width=\wdenoising, height=\hdenoising]{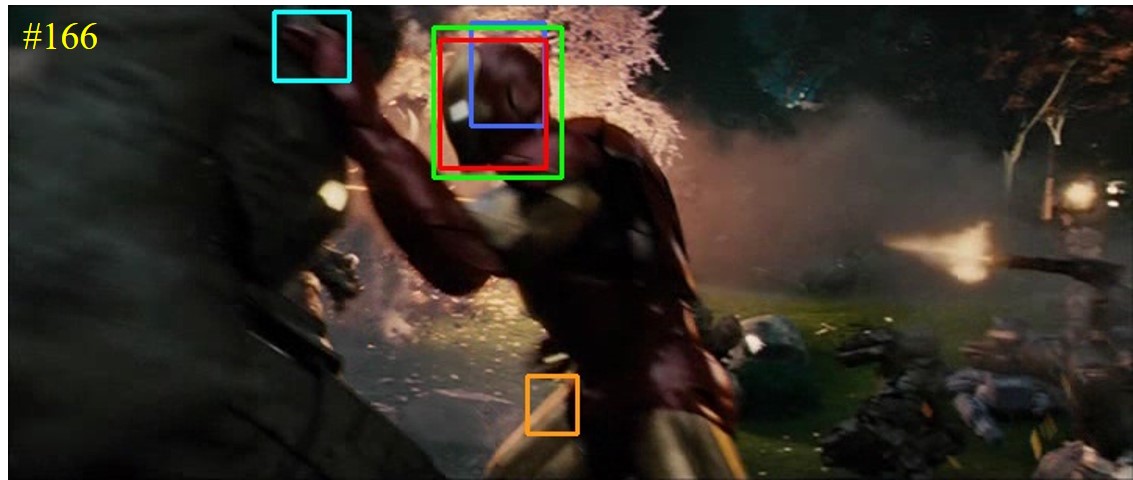} \\
		
		\multicolumn{3}{c}{\includegraphics[width=\wcolor, height=\hcolor]{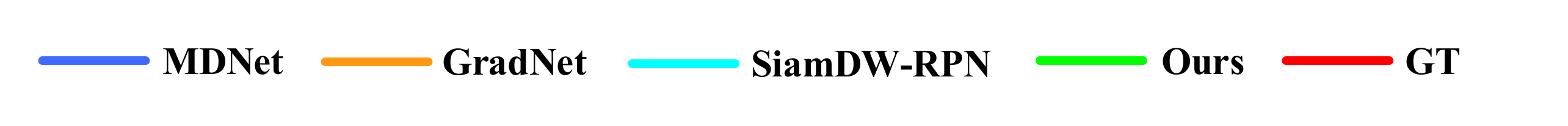}} \\
		\vspace{-0.7cm}
		
	\end{tabular}
	\caption{Visual examples of our tracker comparing with other three state-ot-the-art trackers including GradNet~\cite{li2019gradnet}, MDNet~\cite{nam2016learning}, and SiamDW-RPN~\cite{zhang2019deeper} on \textit{Ironman} from OTB2013~\cite{wu2015object}. \textit{Ironman} is a challenging sequence with low light, motion blur, background clutters, and fast motion. Best viewed in zoom in.}
	\label{fig:teaser}
\end{figure}

Visual object tracking is an important topic in computer vision, where the target object is identified in the first frame and tracked in all frames of a video. 
Due to the significant learning ability, deep convolutional neural networks (DCNNs) have been widely used to object detection~\cite{mei2020don,mei2021exploring,yang2019my}, image matting~\cite{qiao2020multi,qiao2020attention,yang2018active}, super-resolution~\cite{zhang2020multi,zhang2021two,yang2018drfn}, image enhancement~\cite{xu2020learning,yang2018image} and visual object tracking~\cite{bertinetto2016fully,fan2019siamese,he2018twofold,li2019siamrpn++,li2019gradnet,zhang2019deeper,nam2016learning,danelljan2017eco,zhang2017correlation,zhang2018learning,jung2018real,danelljan2019atom,dai2019visual,li2020progressive,wang2017greedy,ren2020tracking}.
However, RGB-based trackers suffer from bad environmental conditions, \textit{e.g.}, low illumination, fast motion, and so on.
Some works~\cite{song2013tracking,kart2018make,kart2018depth,li2018cross,li2019multi,zhu2019dense} try to introduce additional information (\textit{e.g.}, depth and thermal infrared) to improve tracking performance. 
However, when the tracking target is in a high-speed motion or an environment with a wide dynamic range, these sensors usually cannot provide satisfactory results.


\zjq{Event-based cameras are bio-inspired vision sensors whose working principle is entirely different from traditional cameras. 
While conventional cameras obtain intensity frames at a fixed rate, event-based cameras measure light intensity changes and output events asynchronously.}
\zjq{Compared with conventional cameras, event-based cameras have several advantages. First, with a high temporal resolution (around 1 $\mu$$s$), event-based cameras do not suffer from motion blur. Second, event-based cameras have a high dynamic range (\textit{i.e.}, 120-140 dB). Thus they can work effectively even under over/under-exposure conditions. }

We observe that events and RGB data are captured from different types of sensors, but they share some similar information like target boundaries. At the same time, stacked event images and RGB images have their own unique characteristics. In particular, RGB images contain rich low- and high-frequency texture information and provide abundant representations for describing target objects. Events can provide target edge cues that are not influenced by motion blur and bad illumination. Therefore, event-based data and RGB images are complementary, which calls for the development of novel algorithms capable of combining the specific advantages of both domains to perform computer vision in degraded conditions.

To the best of our knowledge, we are the first to jointly explore RGB and events for object tracking based on their similarities and differences in an end-to-end manner. This work is essentially object tracking with multi-modal data that includes RGB-D tracking~\cite{song2013tracking,kart2018make,xiao2017robust,kart2018depth}, RGB-T tracking~\cite{li2018cross,lan2018robust,li2019multi,zhu2019dense,zhang2019multi}, and so on. However, since the output of an event-based camera is an asynchronous stream of events,  this makes event-based data fundamentally different from other sensors' data that have been addressed well by multi-model tracking methods. With the promise of increased computational ability and low power computation using neuromorphic hardware, Spiking Neural Networks (SNNs), a processing model aiming to improve the biological realism of artificial neural networks, show their potential as computational engines, especially for processing event-based data from neuromorphic sensors. Therefore, combining SNNs and DCNNs to process multi-domain data is worth exploring.

In this paper, focusing on the above two points, we propose Multi-domain Collaborative Feature Representation (MCFR) that can effectively extract the common features and unique features from both domains for robust visual object tracking in challenging conditions. 
Specifically, we employ the first three convolutional layers of VGGNet-M~\cite{simonyan2014very} as our Common Features Extractor (CFE) to learn similar potential representations from the RGB domain and event domain. To model specific characteristics of each domain, the Unique Extractor for RGB (UER) is designed to extract unique texture and semantic features in the RGB domain. Furthermore, we leverage the Unique Extractor for Events (UEE) based on SNNs to efficiently extract edge cues in the event domain. Extensive experiments on the RGB benchmark and real event dataset suggest that the proposed tracker achieves outstanding performance. 
A visual example can be seen in Figure~\ref{fig:teaser}, which contains multiple challenging attributes.
By analyzing quantitative results, we provide basic insights and identify the potentials of events in  visual object tracking. 

\zjq{To sum up, our contributions are as follows:}

\zjq{$\bullet$ We propose a novel multi-domain feature representation network which can effectively extract and fuse the information from frame and event domains.}

\zjq{$\bullet$ We preliminarily explore combining SNNs and DCNNs for visual object tracking.}

\zjq{$\bullet$ The extensive experiments verify our approach outperforms other state-of-the-art methods. The ablation studies evidence the effectiveness of the designed components.}
\section{Related Work}

\subsection{\zjq{Spiking Neural Networks}}
Spiking Neural Networks (SNNs) are bio-inspired models using spiking neurons as computational models. The inputs of spiking neurons are temporal events called spikes, and the outputs also are spikes. Spiking neurons have a one-dimensional internal state named potential, which is controlled by first-order dynamics. Whenever a spike arrives, if no other spikes are recorded in time, the potential will be excited but will decay again. When the potential reaches a certain threshold, the spiking neuron sends a spike to the connected neurons and resets its own potential. 
It has been shown that such networks are able to process asynchronous, without pre-processing events data ~\cite{cohen2016skimming,gehrig2019end}. Since the spike generation mechanism cannot be differentiated and the spikes may introduce the problem of incorrect allocation of the time dimension, the traditional gradient backpropagation mechanism cannot be directly used in SNNs. Nonetheless, some researches~\cite{neftci2019surrogate,zenke2018superspike,shrestha2017robustness,shrestha2018slayer,tavanaei2019deep,gehrig2020event} on supervised learning for SNNs has taken inspiration from backpropagation to solve the error assignment problem. 
However, it is still unclear how to train multiple layers of SNNs, and combine them with DCNNs for tracking task.

\subsection{\zjq{Single-Domain Tracking}}

\textbf{RGB-based tracking.} \zjq{Deep-learning-based methods have dominated the visual object tracking field, from the perspective of either one-shot learning~\cite{bertinetto2016fully,fan2019siamese,he2018twofold,li2019siamrpn++,li2019gradnet,zhang2019deeper} or online learning~\cite{nam2016learning,danelljan2017eco,zhang2017correlation,zhang2018learning,jung2018real,danelljan2019atom,dai2019visual,li2020progressive}.} Usually, the latter methods are more accurate (with less training data) but slower than the former  ones. 
Among them, Nam \textit{et al.} ~\cite{nam2016learning} proposed the Multi-Domain Network (MDNet), which used a CNN-based backbone pretrained offline to extract generic target representations, and the fully connected layers updated online to adapt temporal variations of target objects. In MDNet~\cite{nam2016learning}, each domain corresponds to one video sequence. Due to the effectiveness of this operation in visual tracking, we follow this idea to ensure the accuracy of tracking.

\noindent
\textbf{Event-based tracking.} Compared with the frame-based object tracking methods, there are only a few works on event-based object tracking~\cite{piatkowska2012spatiotemporal,mitrokhin2018event,zhu2020event,9292994,barranco2018real,stoffregen2019event,chen2020end}.
Piatkowska \textit{et al.}~\cite{piatkowska2012spatiotemporal} presented a Gaussian mixture model to track the pedestrian motion. Barranco \textit{et al.}~\cite{barranco2018real} proposed a real-time clustering algorithm and used Kalman filters to smooth the trajectories. \zjq{Zhu \textit{et al.}~\cite{zhu2020event} monitored the confidence of the velocity estimate and triggered a tracking command once the confidence reaches a certain threshold.
Ramesh \textit{et al.}~\cite{9292994} presented a long-term object tracking framework with a moving event camera under general tracking conditions.}
Mitrokhin \textit{et al.}~\cite{mitrokhin2018event} proposed a motion compensation method for tracking objects by getting the possible areas that are not consistent with camera motion. Timo.S \textit{et al.} ~\cite{stoffregen2019event} calculated the optical flow from the events at first, then warped the events' position to get the sharp edge event images according to the contrast principle. Besides, they gave each event a weight as its probability and fused them during the process of warping so that they can classify events into different objects or background. Chen \textit{et al.}~\cite{chen2020end} proposed an end-to-end retinal motion regression network to regress 5-DoF motion features.

Although the above studies have achieved good performance in the RGB domain or the event domain, they ignore exploring the complementary information existing between the two domains. 
As a consequence, we investigate the similarities and differences between the event and RGB domain, and propose common features extractor and unique feature extractor to learn and fuse valuable complementary features.

\subsection{\zjq{Multi-Domain Tracking}}
The current popular visual object tracking based on multi-domain data mainly includes RGB-D (RGB + depth) tracking ~\cite{song2013tracking,xiao2017robust,kart2018depth,kart2018make} and RGB-T (RGB + thermal) tracking~\cite{li2018cross,lan2018robust,li2019multi,zhu2019dense,zhang2019multi}.
Depth cues are usually introduced to solve the occlusion problem in visual object tracking. Images from the thermal infrared sensors are not influenced by illumination variations and shadows, and thus can be combined with RGB to improve performance in bad environmental conditions.
As the output of an event camera is an asynchronous stream of events,  this makes raw event stream fundamentally different from other sensors data that have been addressed well by the above state-of-the-art multi-model tracking methods.
Therefore, it is essential to design a tailored algorithm for leveraging RGB data and event data simultaneously.

\section{Methodology}
\begin{figure*}[t]
	\centering
	\includegraphics[width = 1\linewidth]{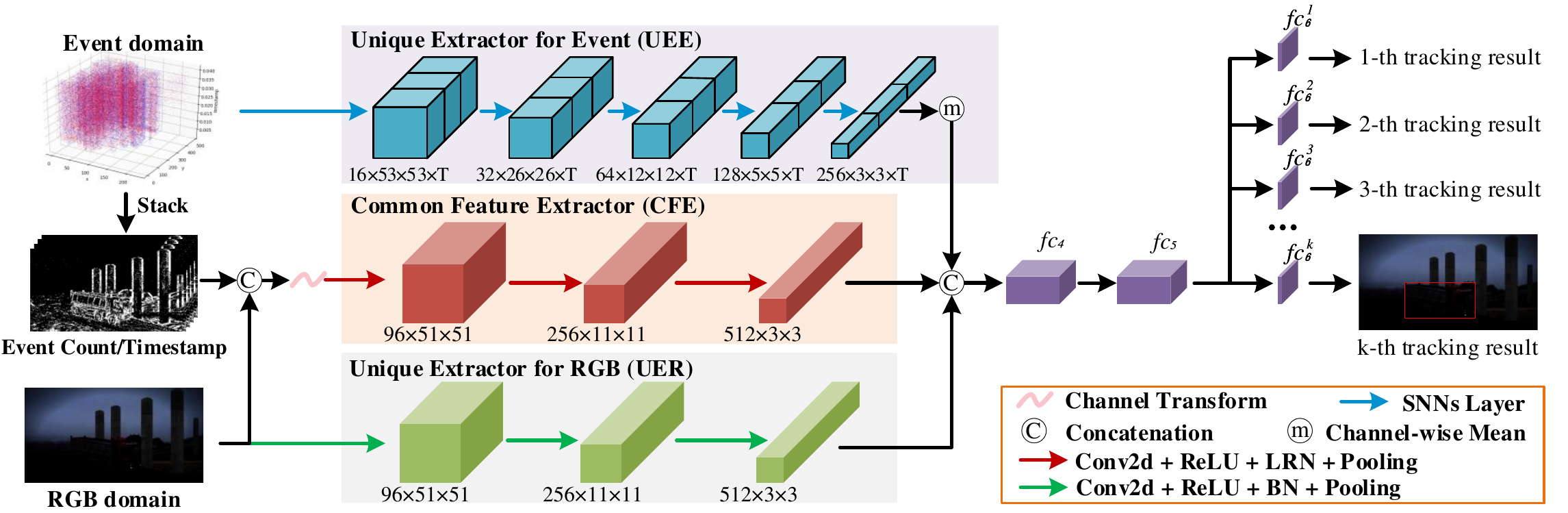}
	\caption{The overview of our proposed network. Our pipeline mainly consists of three parts, UEE for extracting special features from event domain, UER for extracting unique features from RGB domain, and CFE for extracting common shared features from both domains. The target is a moving truck in underexposure.}
	\label{fig:pipeline}
\end{figure*}

\subsection{\zjq{Backgroud: Event-based Camera}}
An event-based camera is a bio-inspired sensor. It asynchronously measures light intensity changes in scene-illumination at a pixel level. Therefore, it provides a very high-temporal resolution (\textit{i.e.}, up to 1MHz). Due the light intensity changes are measured in the log scale, an event-based camera can offer a very high dynamic range (\textit{i.e.}, up to 140 dB). An event was triggered when the change of a log-scale pixel intensity is higher or lower than a threshold, resulting in an ``ON'' and an ``OFF'' event, respectively. Mathematically, a set of events can be defined as:
\begin{equation} 
\label{eq:event_set}
\mathcal{E} = \{e_k\}_{k=1}^N = \{[x_k, y_k, t_k, p_k]\}_{k=1}^N,
\end{equation}
where $e_k$ is the $k$th event; $[x_k, y_k]$ is the pixel location of event $e_k$; $t_k$ is the timestamp when the event is triggered; $p_k \in \{-1, 1\}$ is the polarity of an event, where $- 1$ and $1$ represent OFF and ON events, respectively. In a constant lighting condition, events are normally triggered by moving edges (\textit{e.g.}, object contour, texture and depth discontinuities), which makes an event-based camera be a natural edge extractor. Therefore, with these unique features, event-based cameras have been introduced to various tasks~\cite{stoffregen2019event,tulyakov2019learning,bi2019graph,kepple2020jointly,choi2020learning,pan2020single,cadena2021spade,mostafavi2021learning} in challenging scenes (\textit{e.g.}, low-light, fast motion).

Even though event-based cameras are sensitive to edges, they cannot provide absolute intensity and texture information. Besides, since the asynchronous event stream differs significantly from the frames generated by conventional frame-based cameras, vision algorithms designed for frame-based cameras cannot be directly applied. To deal with it, events are typically aggregated into a grid-based representation first.


\subsection{\zjq{Network Overview}}
Our approach builds on two key observations. First, although events and RGB data are captured from different types of sensors, they share some similar information, such as target object boundaries. Similar features should be extracted using a consistent strategy. Second, rich textural and semantic cues can be easily captured by a conventional frame-based sensor. In contrast, an event-based camera can easily capture edge information which may be missed in RGB images under some challenging conditions. Therefore, fusing complementary advantages of each domain will enhance feature representation.
Figure~\ref{fig:pipeline} illustrates the proposed Multi-domain Collaborative Feature Representation (MCFR) for robust visual object tracking. 
Specifically, for the first observation, we propose the Common Feature Extractor (CFE) which accepts stacked event images and RGB images as inputs to explore shared common features. 
For the second observation, we design a Unique Extractor for Event (UEE)  based on SNNs to extract edge cues in the event domain which may be missed in the RGB domain under some challenging conditions, and a Unique Extractor for RGB (UER)  based on DCNNs to extract texture and semantic information in the RGB domain.
The outputs of UEE, CFE, and UER are then concatenated, and a convolutional layer with 1$\times$1 kernel size is used to adaptively select valuable combinative features.  Finally, the combinative features are classified by three fully connected layers and softmax cross-entropy loss. Following~\cite{nam2016learning}, the network has $k$ branches, which are denoted by the last fully connected layers. In other words, training sequences $fc6^1 - fc6^k$.

\subsection{\zjq{Common Feature Extractor}}
To leverage a consistent scheme for extracting similar features of event and RGB domains, we first stack event stream according to the counts and latest timestamp of positive and negative polarities, which makes vision algorithms designed for frames can also be applied to asynchronous event streams. Mathematically,
\begin{equation}
\begin{aligned}
C(x, y, p)&=\sum_{k=1, t_{k} \in W}^{N} \delta\left(x_{k}, x\right) \delta\left(y_{k}, y\right) \delta\left(p_{k}, p\right) \\
T(x, y, p)&=\max _{t_{k} \epsilon W} t_{k} \delta\left(x_{k}, x\right) \delta\left(y_{k}, y\right) \delta\left(p_{k}, p\right)
\end{aligned}
\end{equation}
where $\delta$ is the Kronecker delta function, $W$ is the time window (the interval between adjacent RGB frames), and $N$ is the number of events that occurred within $W$.
The stacked event count image $C$ contains the number of events at each pixel, which implies the frequency and density information of targets. The stacked event timestamp image $T$ contains the temporal cues of the motion, which implies the direction and speed information of targets. 
An example of counts images and timestamp images is shown in Figure~\ref{fig:counttime}, we find that the stacked event images and RGB image indeed share some common features, such as the edge cues of targets. 

We then employ a Common Feature Extractor (CFE) to extract shared object representations across different domains. To balance effectiveness and efficiency, we apply the first three layers from the VGGNet-M~\cite{simonyan2014very} as the main feature extraction structure of our CFE. Specifically, the convolution kernel sizes are $7\times7$, $5\times5$, and $3\times3$, respectively. The output channels are 96, 256, and 512, respectively. As shown in Figure~\ref{fig:pipeline}, 
the whole process is formulated as 
$F_{cfe} = CFE( \tau([RGB, C, T]))$,
where $RGB$ denotes RGB image, $[\cdot]$ is concatenation, $\tau$ indicates channel transformation, and $F_{cfe}$ is the output of CFE.

\def\wdenoising{0.19\linewidth}
\def\hdenoising{0.8in}
\begin{figure*}[t]
	\setlength{\tabcolsep}{2.4pt}
	\centering
	\begin{tabular}{ccccc}
		
		\includegraphics[width=\wdenoising, height=\hdenoising]{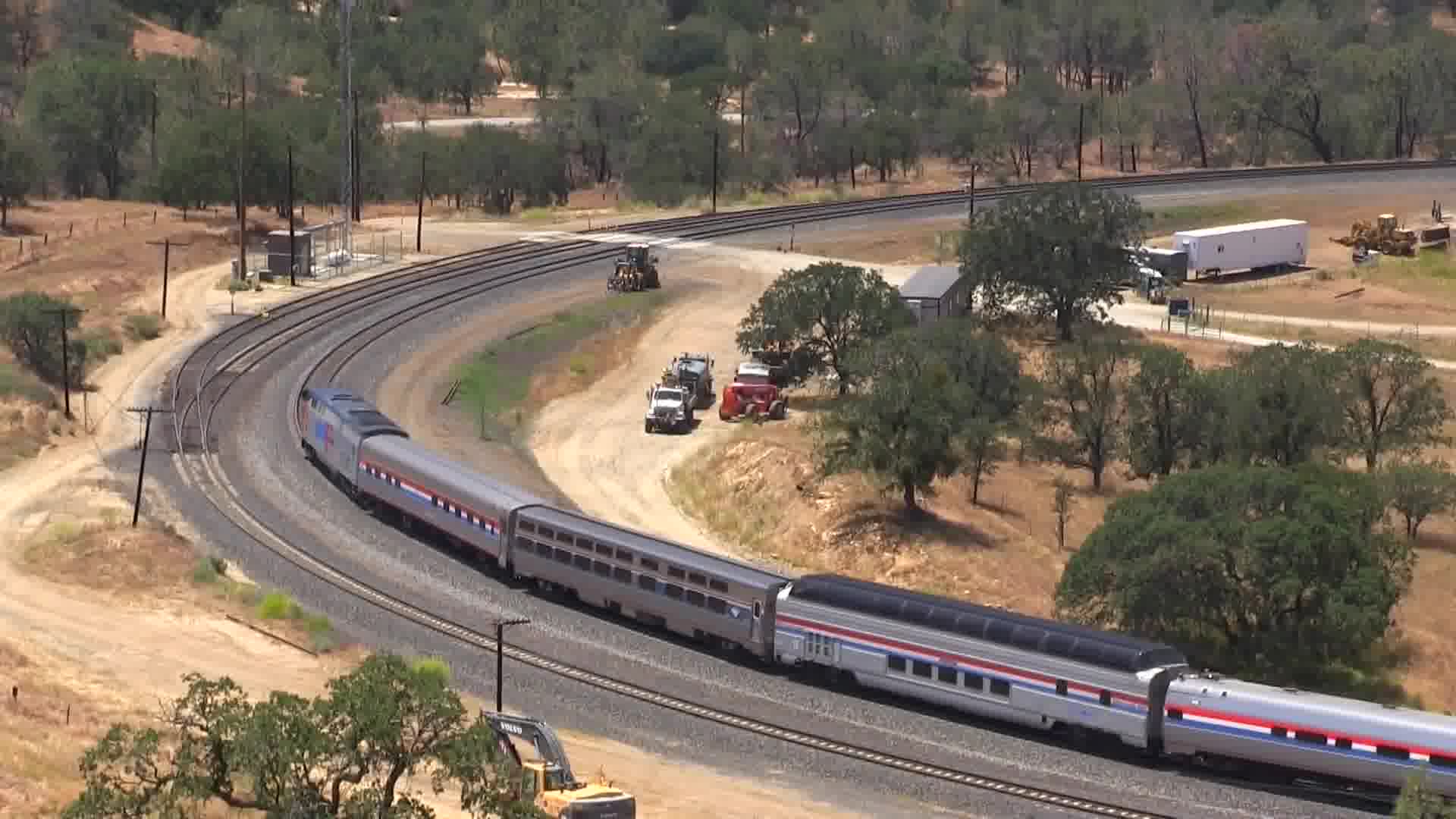}&
		\includegraphics[width=\wdenoising, height=\hdenoising]{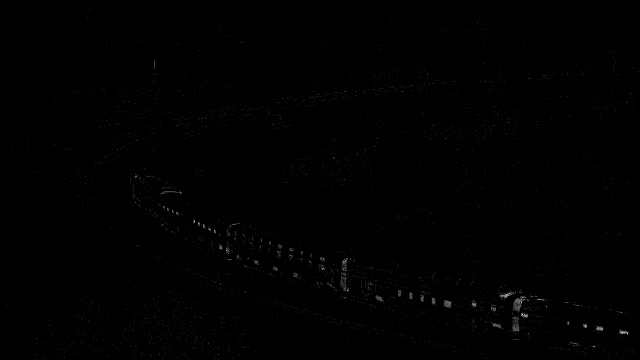}&
		\includegraphics[width=\wdenoising, height=\hdenoising]{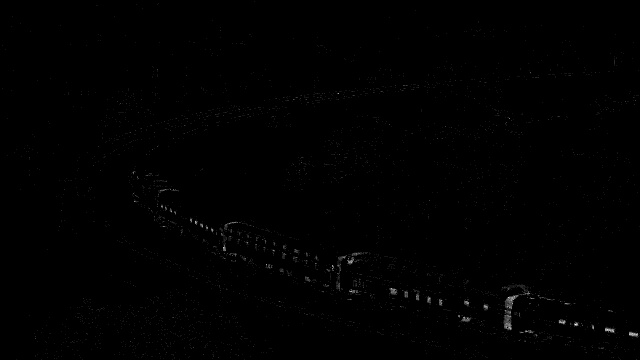}&
		\includegraphics[width=\wdenoising, height=\hdenoising]{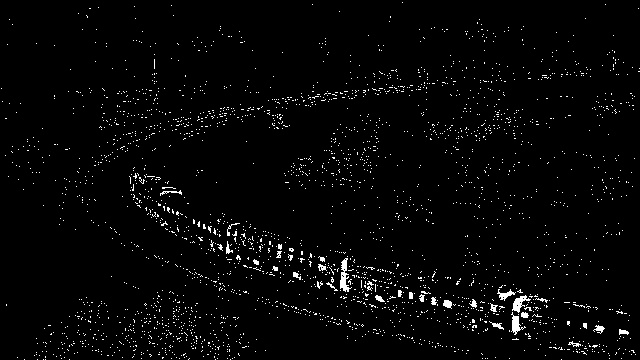}&
		\includegraphics[width=\wdenoising, height=\hdenoising]{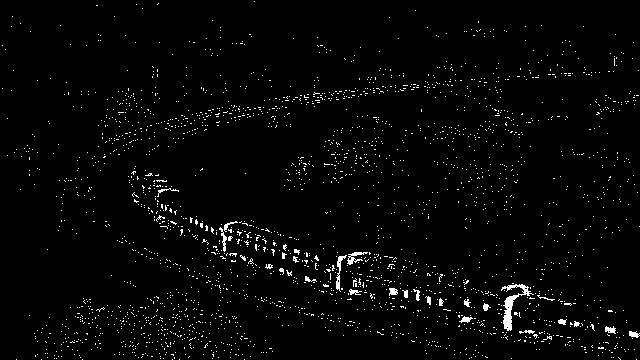}\\
		
		(a) RGB & (b)pos\_counts & (c) neg\_counts & (d) pos\_timestamp & (e) neg\_timestamp\\
		
	\end{tabular}
	\caption{Example of counts and timestamp images. Left to right: RGB image, positive counts image, negative counts image, positive timestamp image, and negative timestamp image. In timestamp images, each pixel represents the timestamp of the most recent event, and brighter is more recent.}
	\label{fig:counttime}
\end{figure*}

\subsection{\zjq{Unique Extractor for RGB}}
Since the raw event stream and RGB data storage methods and expressions are different, it is necessary to design an exclusive feature extraction method for each domain. For the RGB domain, we propose Unique Extractor for RGB (UER) to effectively extract unique texture and semantic features.
Specifically, as shown in Figure~\ref{fig:pipeline}, UER consists of three convolutional layers, and the size of the convolution kernel are $3\times3$, $1\times1$, and $1\times1$, respectively. 
It is noted that one major difference between UER and CFE is the size of the convolution kernel. CFE employs large-size convolution kernels to provide a larger receptive field so that the whole boundary from RGB and event domains can be better extracted, while UER can focus on the rich texture information in the RGB domain with small-size kernels.
This process can be simply formulated as $F_{uer} = UER(RGB)$,
where $F_{uer}$ is the output of UER.

%
%

\subsection{\zjq{Unique Extractor for Event}}
Compared with RGB images, the event-based data is not affected by HDR and motion blur. Besides, from Figure~\ref{fig:counttime}, we can see that events can provide clear cues about where object movement occurred, which will help the tracking process not be disturbed by the surrounding environment. 
Since SNNs can process raw event stream directly, we introduce it into our Unique Extractor for Events (UEE) (top branch in Figure~\ref{fig:pipeline}) to effectively extract unique event features. 
There are different mathematical models to describe the dynamics of a spiking neuron, we use the Spike Response Model (SRM)~\cite{gerstner1995time} in this work. 
In the SRM~\cite{gerstner1995time},  the net effect that firing has on the emitting and the receiving neuron is described by two response functions, $v(t)$ and $u(t)$. The refractory function $u(t)$ describes the response of the firing neuron to its own spike. The synaptic kernel $v(t)$ describes the effect of an incoming spike on the membrane potential at the soma of the postsynaptic neuron.
Following ~\cite{gehrig2020event,Shrestha2018}, we define the feedforward SNNs with $n$ layers as:
\begin{equation}
v(t) = \frac{t}{\tau_{s}} e^{1-\frac{t}{\tau_{s}}} H(t), \quad u(t) = -2\phi e^{-\frac{t}{\tau_{r}}} H(t)
\end{equation}
\vspace{-0.45cm}
\begin{equation}
\varepsilon_{i+1}(t) = W_i(v * s_i)(t) + (u * s_{i+1})(t)
\end{equation}
\vspace{-0.45cm}
\begin{equation}
s_i(t) = \sum\delta(t - t_i) \quad (t_i\in \{t | \varepsilon_i(t) = \phi\})
\end{equation}
\vspace{-0.45cm}
\begin{equation}
F_{uee} = \mathcal{M}(W_n(v * s_n)(t))
\end{equation}
where $H$ is the Heaviside step function; $\tau_{s}$ and $\tau_{r}$ are the time constants of the synaptic kernel and refractory kernel, respectively.
$s_i$ and $W_i$ are the input spikes and synaptic weights of the $i$th layer, respectively.  $\phi$ denotes the neuron threshold, that means, when the sub-threshold membrane potential is strong enough to exceed $\phi$ the spiking neuron responds with a spike. To combine SNNs with DCNNs in the overall structure, we perform a mean operation $\mathcal{M}$ on the time dimension \emph{T} of SNNs output. $F_{uee}$ is the output of our UEE. 

\subsection{\zjq{Discussion}}
After extracting common shared features and unique features from both domains, we fuse them with a concatenate operation. 
Considering different video sequences have different classes, movement styles, and challenging aspects,
we further use three fully connected layers named as $fc_4$,  $fc_5$, and  $fc_6$ whose output channels are 512, 512, and 2, respectively, to further process fusion features. $fc_6$ is a domain-specific layer, that means each training has $k$ sequences, then there are $k$ $fc_6$ layers. Each of the $k$ sequences contains a binary classification layer with softmax cross-entropy loss, which is responsible for distinguishing target and background. 

It should be noted that we did not use a very deep network or complex integration strategy because of the following reasons. 
First, compared with visual recognition problems, visual tracking requires much lower model complexity because it aims to distinguish only two categories of target and background.
Second, since the target is usually small, it is desirable to reduce the input size, which will naturally reduce the depth of the network.
Finally, due to the need for online training and testing, a smaller network will be more effective.
Our main principle of network design is to make it simple yet work. To the best of our knowledge, this work is the first to explore and utilize the correlation between RGB images and event-based data for visual object tracking. We believe that more and more related works could be done to further improve such a compact network.
    
\subsection{\zjq{Training Details}}
For CFE, we initialize parameters of it using the pre-trained model in VGGNet-M~\cite{simonyan2014very}. For UEE, by the public SLAYER~\cite{Shrestha2018}, we can calculate the gradient of the loss function relative to the SNNs parameter based on the first-order optimization method. We initialize parameters of UEE using the pre-trained model in~\cite{gehrig2020event} and then fix them. 
We use the stochastic gradient descent algorithm (SGD)  to train the network. The batch size is set to 8 frames which are randomly selected from training video sequences. We choose 32 positive samples (IoU overlap ratios with the ground truth bounding box are larger than 0.7) and 96 negative samples (IoU overlap ratios with the ground truth bounding box are less than 0.5) from each frame,  which results in 256 positive and 768 negative samples altogether in a mini-batch. 
For multi-domain learning with $k$ training sequences, we train the network by softmax cross-entropy loss. The learning rates of all convolutional layers are set to 0.0001,  the learning rates of \emph{fc4} and \emph{fc5} are set to 0.0001, and the learning rate of \emph{fc6} is set to 0.001.
 
\subsection{\zjq{Tracker Details}}
During the tracking process, for each test video sequence, we replace \emph{k} branches of \emph{fc6} with a single branch. To capture the context of a new sequence and learn video-specific information adaptively, we adopt online fine-tuning. Specifically, we fix all convolutional filters of UEE, CFE, and UER, and fine-tune the fully connected layers \emph{fc4}, \emph{fc5}, and a single branch \emph{fc6}. The reason is that convolutional layers can extract the generic information about tracking, while the fully connected layers are able to learn video-specific information. For online updating, we collect 500 positive samples (IoU overlap ratios with the ground truth bounding box are greater than 0.7) and 5000 negative samples (IoU overlap ratios with the ground truth bounding box are less than 0.5) as the training samples in the first frame. For the $t$-th frame, we collect a set of candidate regions $z^i_t$ from previous tracking result $Z_{t-1}$ by Gaussian sampling. We then use these candidates as inputs to our network and obtain their classification scores. The positive and negative scores are computed using the trained network as $f^+(z^i_t)$ and $f^{-}(z^i_t)$, respectively. We select the candidate region with the highest score as the target location $Z^{*}_{t}$ of the current frame:
\begin{equation}
Z^{*}_{t} = \arg\max_{z^i_t}f^+(z^i_t), \quad i = 1,2, ...,N
\end{equation}
where $N$ is the number of candidate regions. We use the bounding box regression technique to improve the problem of target scale
transformation in the tracking process and improve the accuracy of positioning.
\section{Experiments}

\subsection{\zjq{Training Dataset Generation}}
Supervised learning for visual object tracking requires a large quantity of data. \zjq{In our case, we need a dataset that contains RGB data from a traditional APS camera (an APS (Active Pixel Sensor) is a conventional image sensor where each pixel sensor unit cell has a photodetector and one or more active transistors) and events from an event-based camera with ground truth bounding box.} Our data set needs to meet the following needs: First, the RGB images and event-based data must be aimed at the same scene, and the data between different domains must be aligned. Second, we must have a large variety of scenes with ground truth bounding boxes to avoid overfitting to specific visual patterns. To our knowledge, such data sets do not yet exist. In order to meet the above requirements, we generate a synthetic dataset using event-camera simulator ESIM~\cite{rebecq2018esim} on large-scale short-term generic object tracking database GOT-10k~\cite{huang2019got}. ESIM~\cite{rebecq2018esim} has successfully been proven its effectiveness in previous works~\cite{wang2019event,rebecq2019events,stoffregen2019event}. GOT-10k~\cite{huang2019got} is a large, high-diversity, and one-shot tracking database with a wide coverage of real-world moving objects. GOT-10k~\cite{huang2019got} collects over 10,000 videos of 563 object classes and annotates 1.5 million tight bounding boxes manually.

Actually, as we all know, traditional RGB frames suffer from motion blur under fast motion, and also have limited dynamic range resulting in the loss of details. Therefore, directly using the RGB and event pairs from ESIM~\cite{rebecq2018esim} is not an ideal way for training the network, as our goal is to  fully exploit the advantages of event cameras.
\def\wdenoising{0.48\linewidth} 
\def\hdenoising{1.4in}
\begin{figure}[t]
	\setlength{\tabcolsep}{2.4pt}
	\centering
	\begin{tabular}{cc}
		
		\includegraphics[width=\wdenoising, height=\hdenoising]{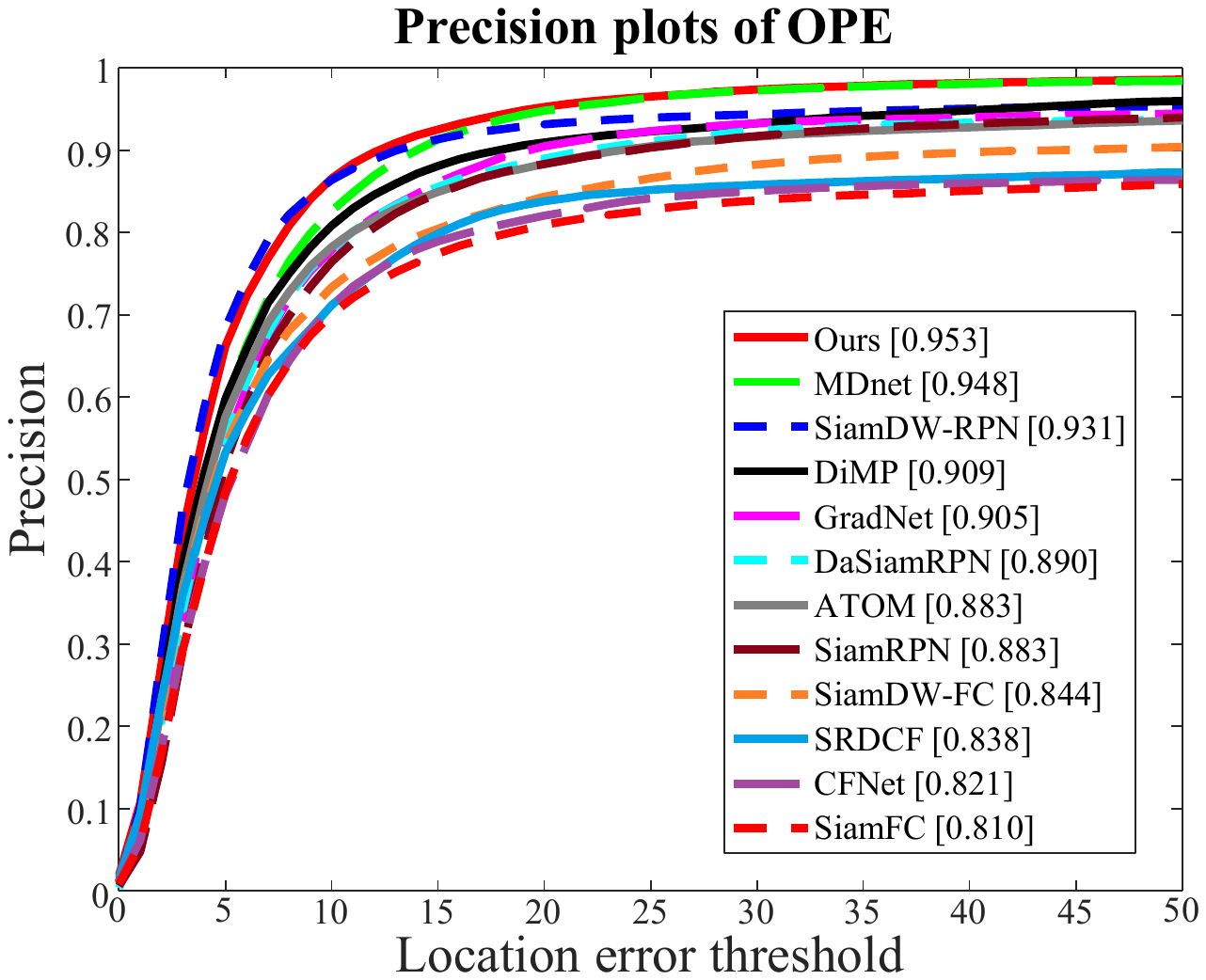}&
		\includegraphics[width=\wdenoising, height=\hdenoising]{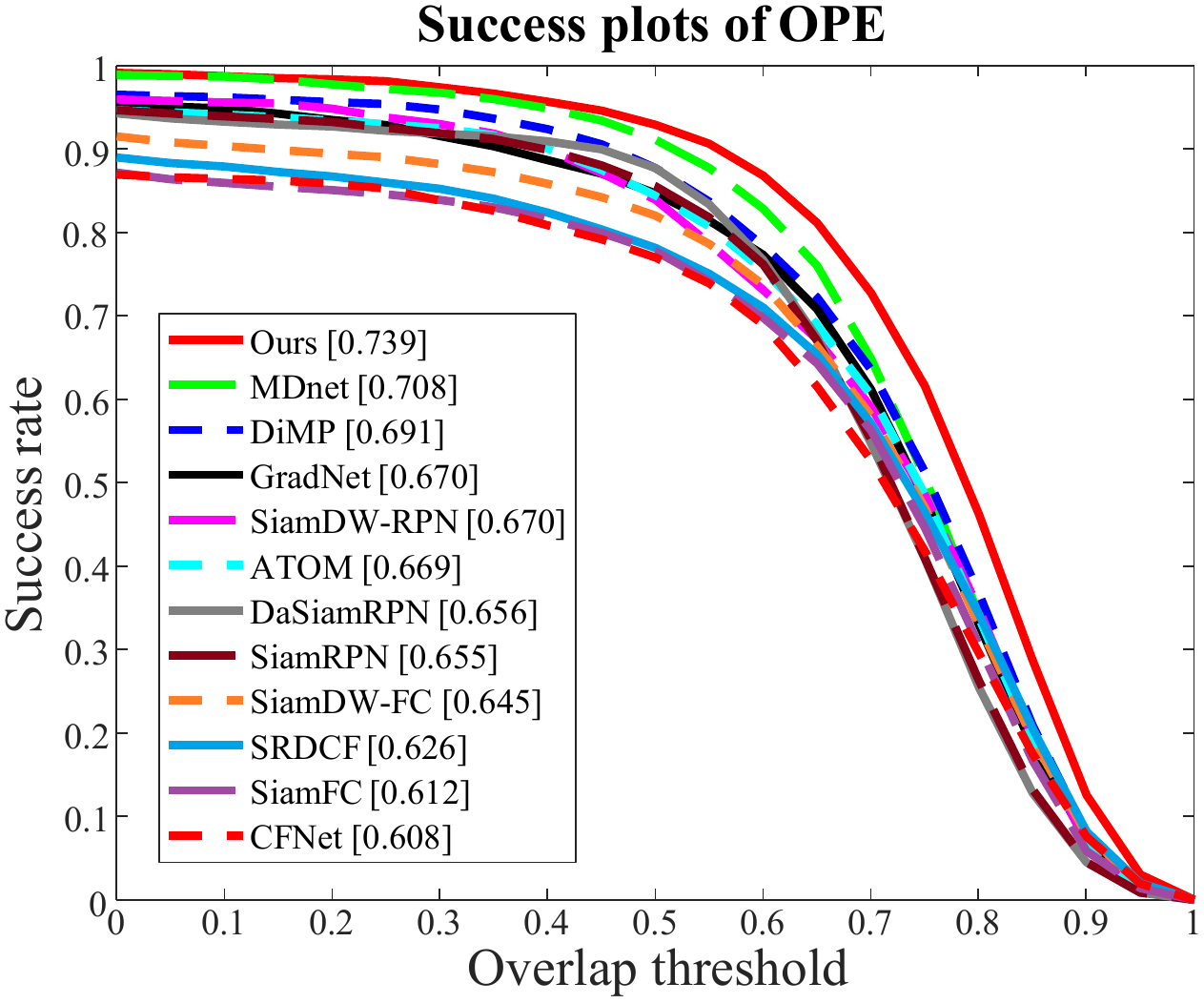} \\
		
	\end{tabular}
	\caption{\zjq{PR and SR curves of different tracking result on OTB2013~\cite{wu2015object} dataset, where the representative PR and SR scores are presented in the legend.}}
	\label{fig:prsr}
\end{figure}
Therefore, we randomly select 100 video sequences. For each RGB frame in the sequence, we randomly increase or decrease the exposure manually. In this way, we simulate the fact that event-based cameras can provide valuable information that conventional cameras cannot capture in extreme exposure conditions.
To verify the effectiveness of our proposed approach, we evaluate it on the standard RGB benchmark and the real event dataset, respectively.

\def\wvisualx2{0.24\linewidth}
\def\hvisualx2{1.4in}
\begin{figure*}[t]
	\setlength{\tabcolsep}{3.0pt}
	\centering
	\begin{tabular}{cccc}
		
		\includegraphics[width=\wvisualx2, height=\hvisualx2]{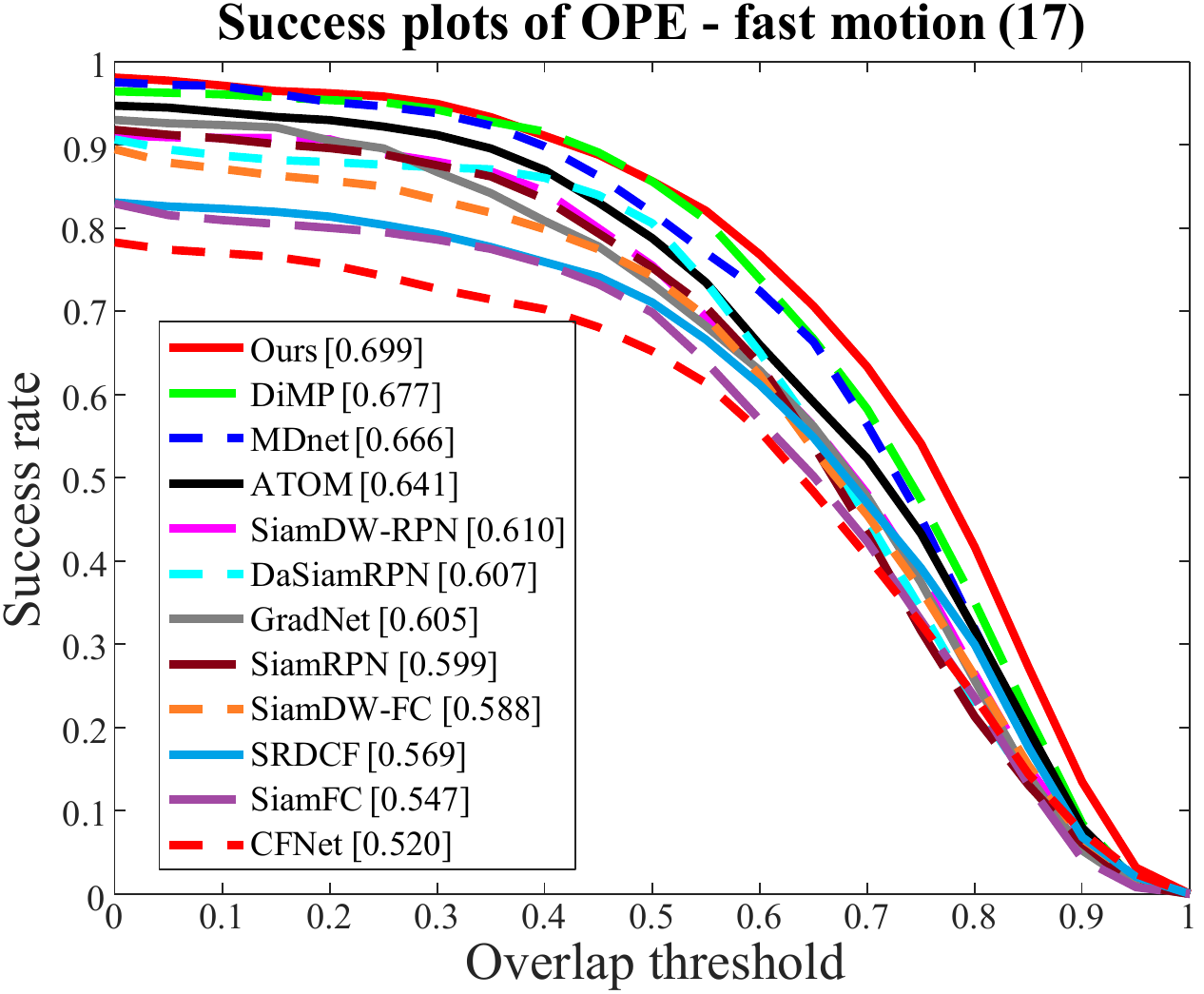}&
		\includegraphics[width=\wvisualx2, height=\hvisualx2]{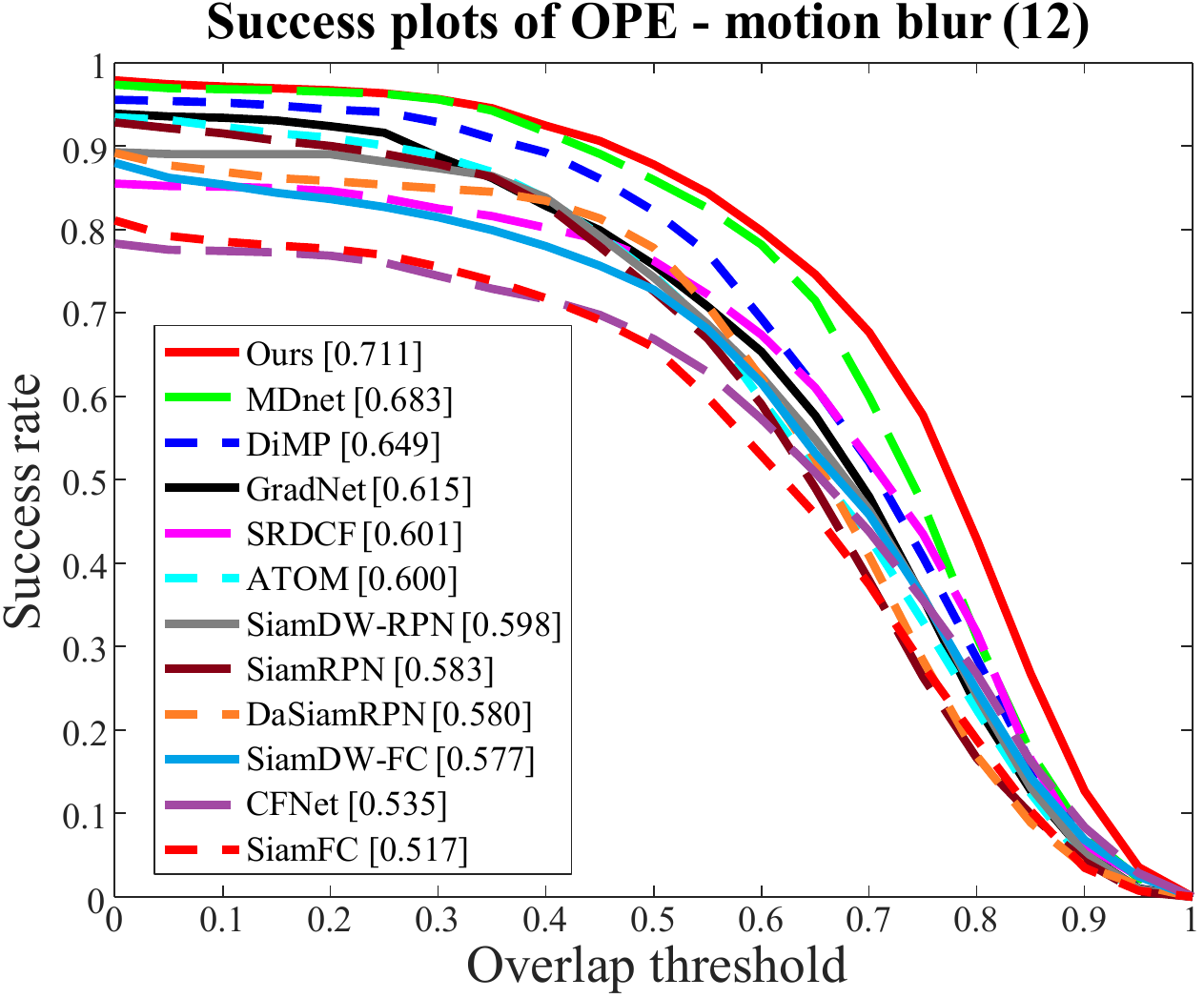}&
		\includegraphics[width=\wvisualx2, height=\hvisualx2]{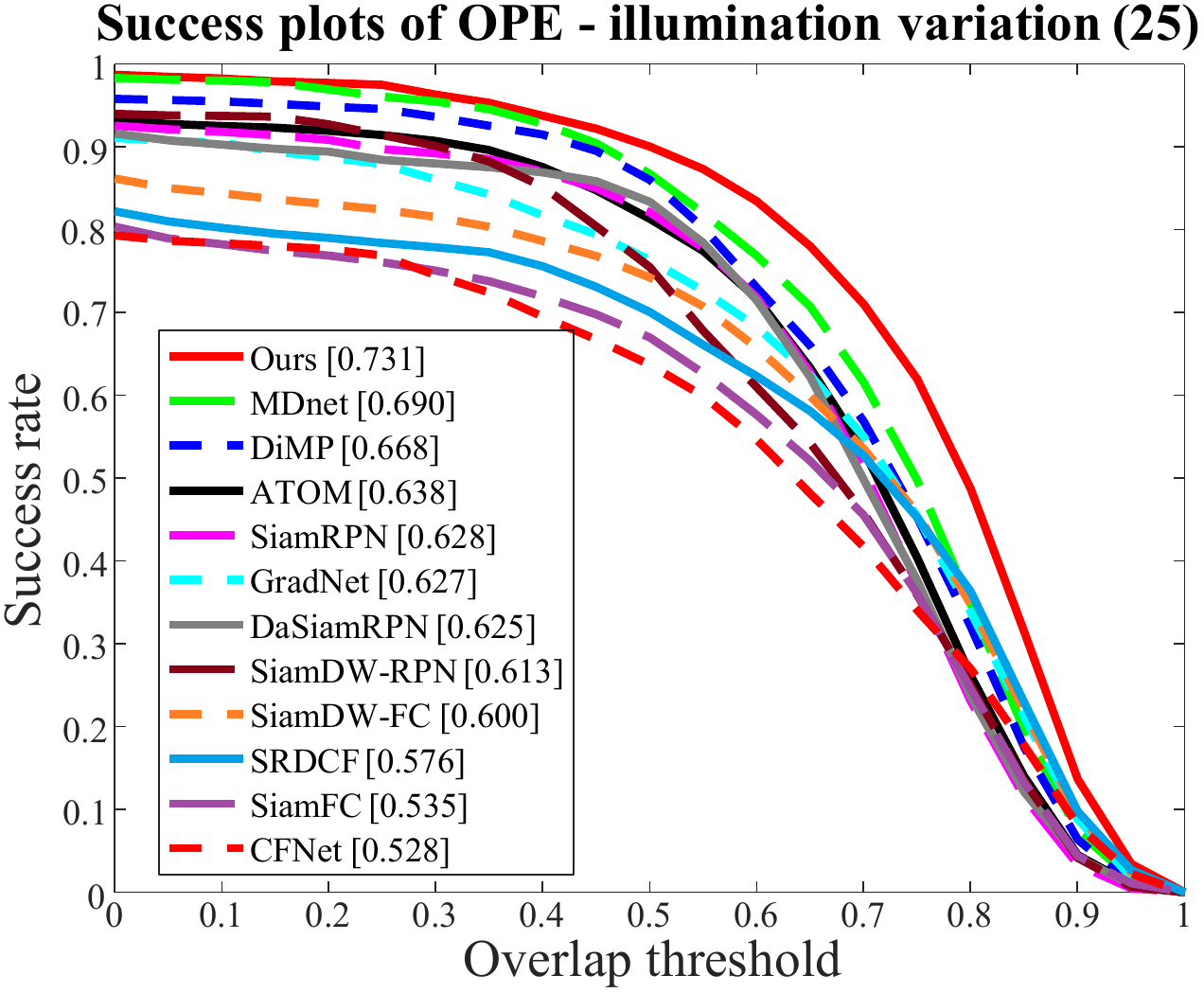}
		&   \includegraphics[width=\wvisualx2, height=\hvisualx2]{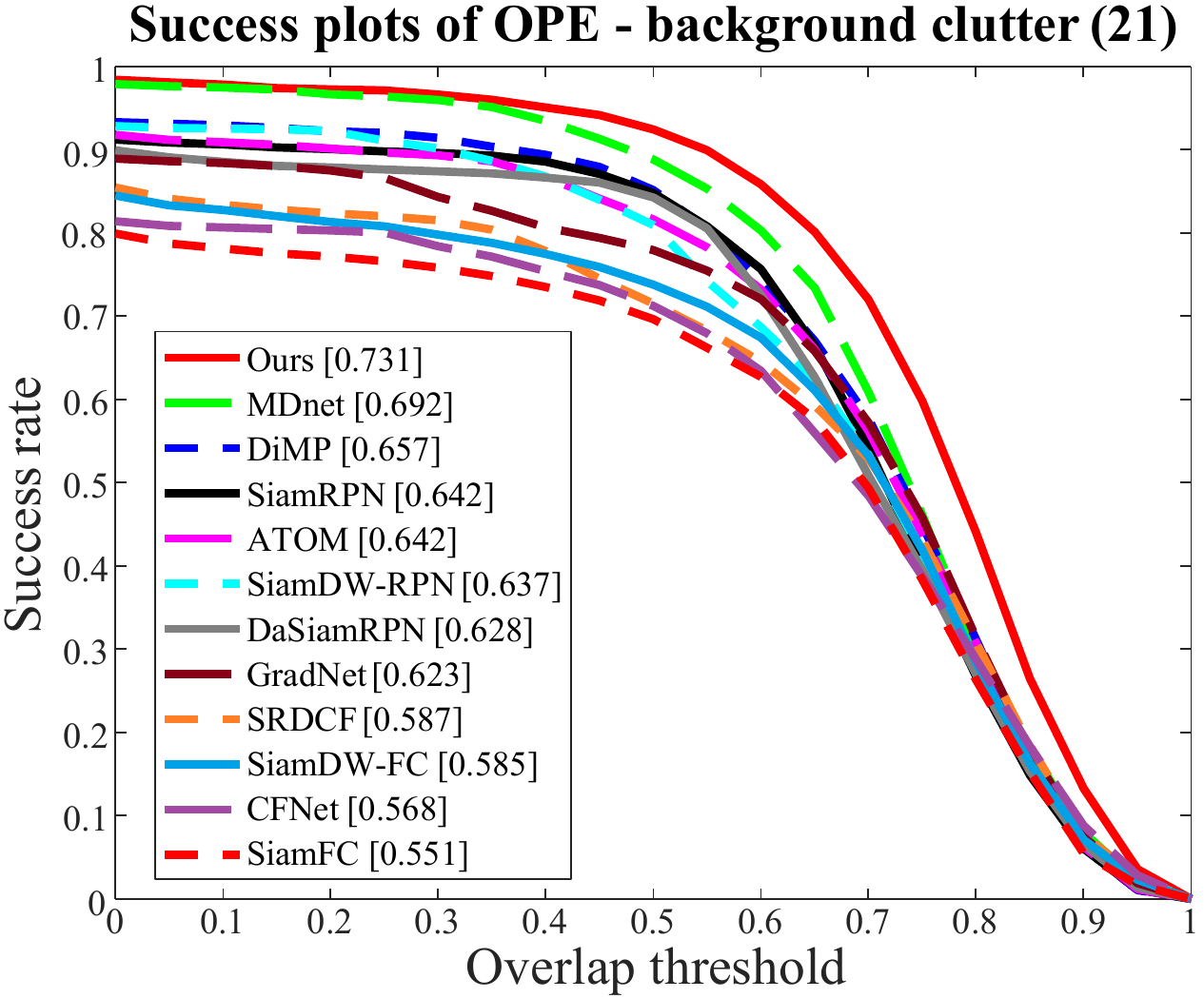} \\	
	\end{tabular}
	\caption{\zjq{Evaluation results on various challenges comparing to the-state-of-the-art methods on OTB2013~\cite{wu2015object}. Left to right: \textit{fast\_motion}, \textit{motion\_blur}, \textit{illumination\_variations} and \textit{background\_clutter}.}}
	\label{fig:OTB2013_various}
\end{figure*}

\def\wdenoising{0.24\linewidth}
\def\hdenoising{0.90in}
\def\wcolor{0.95\linewidth}
\def\hcolor{0.35in}
\begin{figure*}[t]
	\setlength{\tabcolsep}{1.4pt}
	\centering
	\begin{tabular}{ccccc}
		
		\includegraphics[width=\wdenoising, height=\hdenoising]{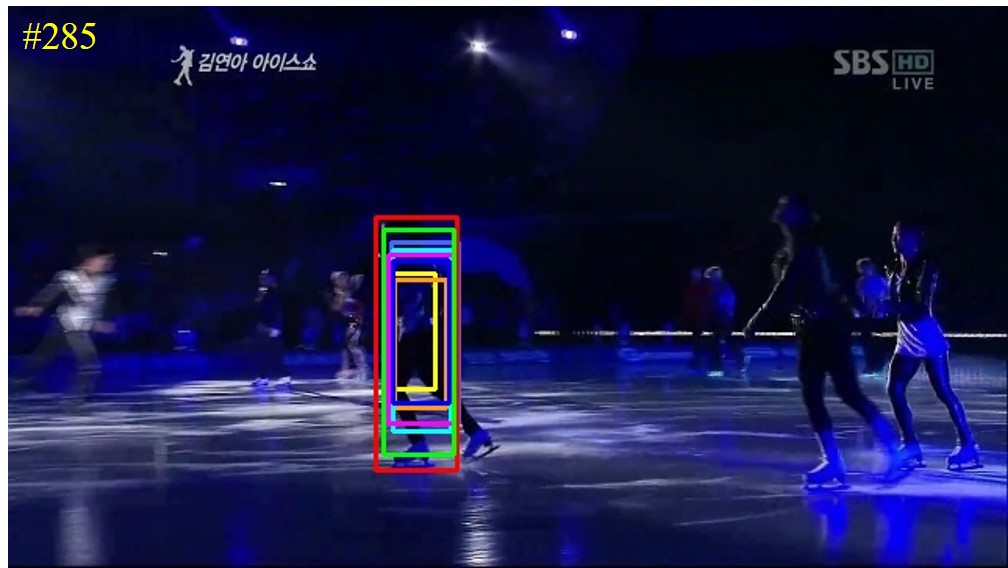}&
		\includegraphics[width=\wdenoising, height=\hdenoising]{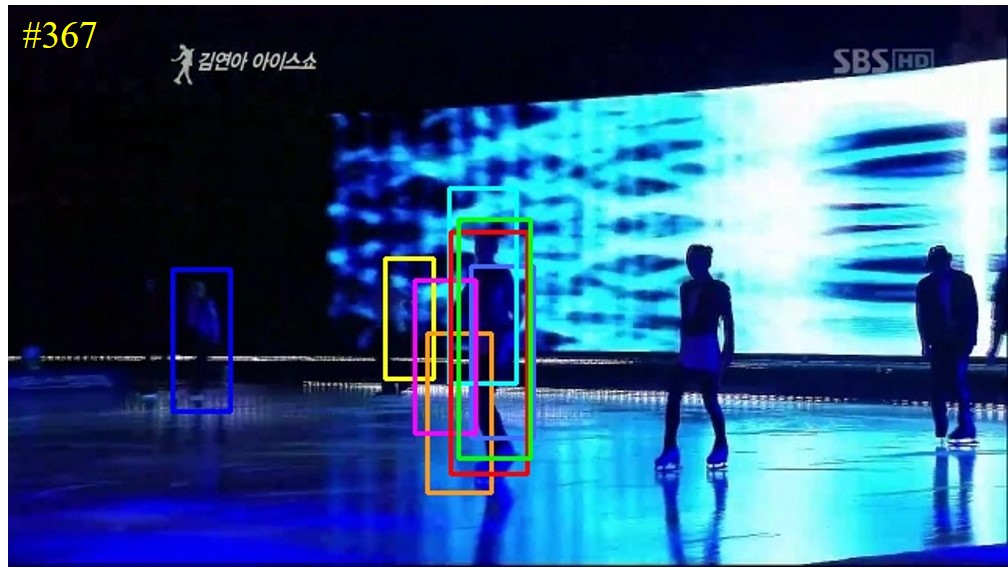}&
		\qquad  &
		\includegraphics[width=\wdenoising, height=\hdenoising]{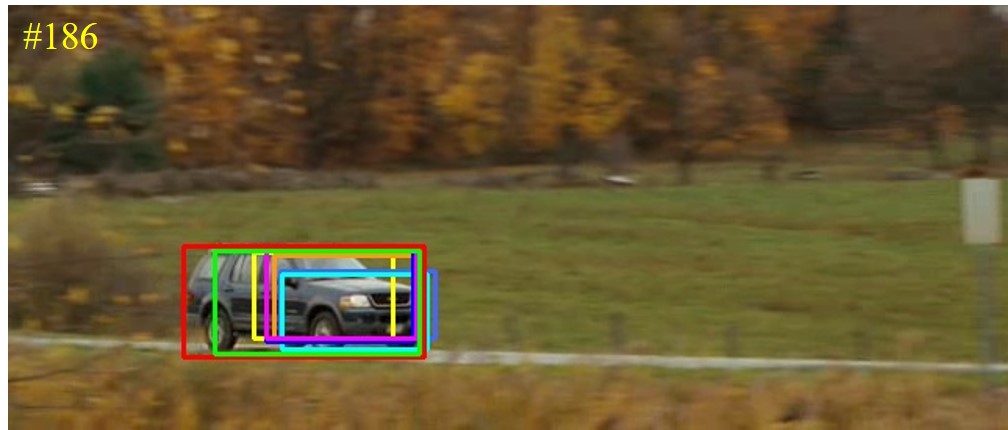}&
		\includegraphics[width=\wdenoising, height=\hdenoising]{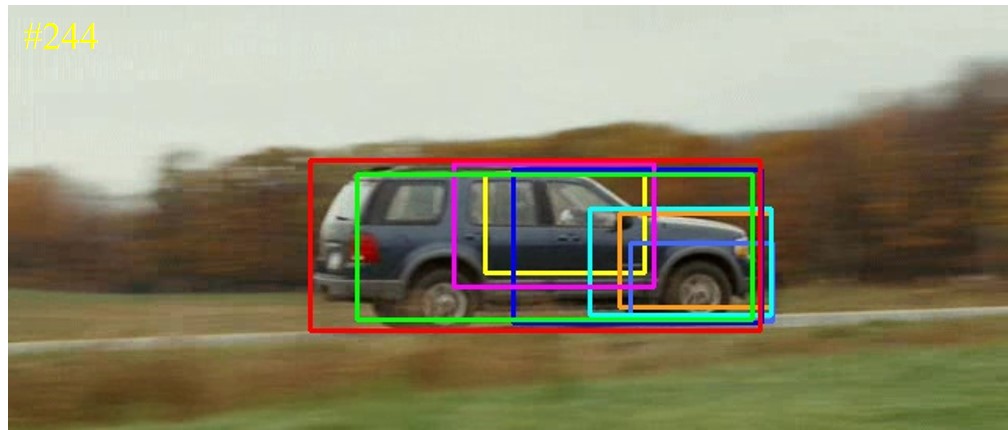}\\
		
		\includegraphics[width=\wdenoising, height=\hdenoising]{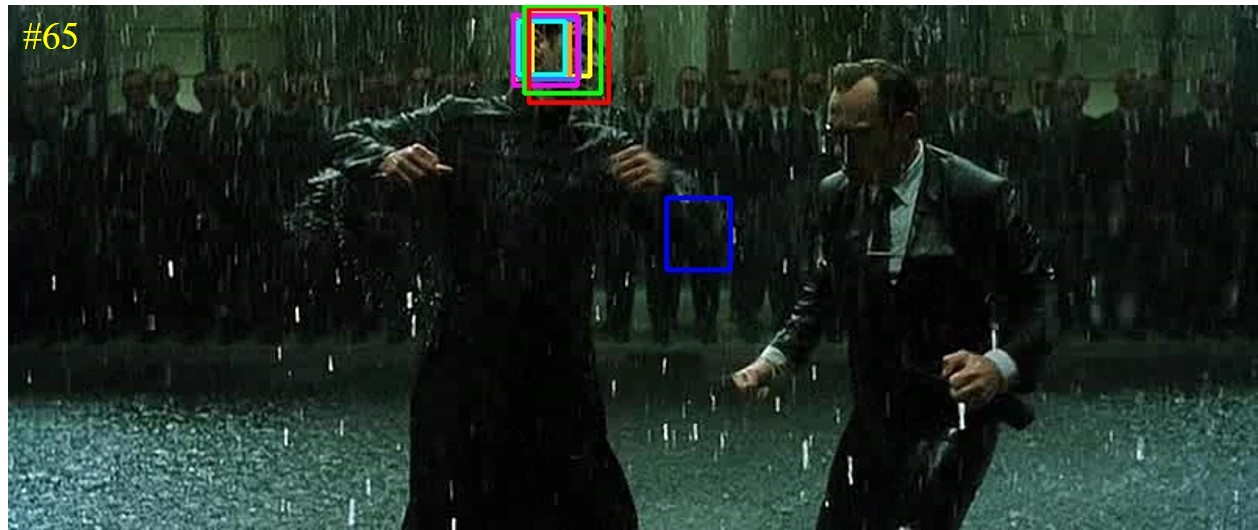}&
		\includegraphics[width=\wdenoising, height=\hdenoising]{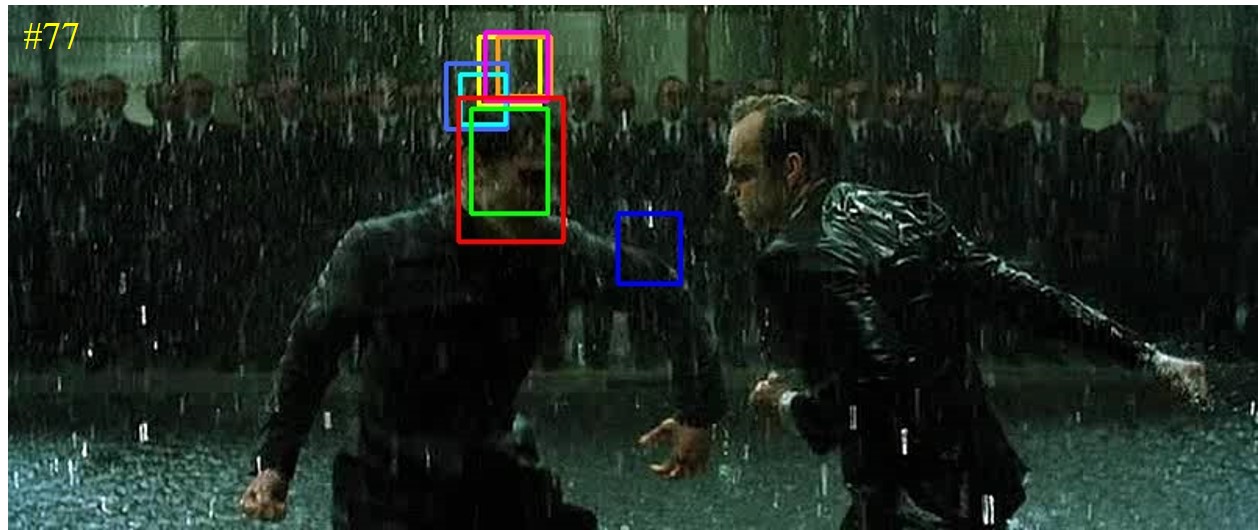}&
		\qquad   &
		\includegraphics[width=\wdenoising, height=\hdenoising]{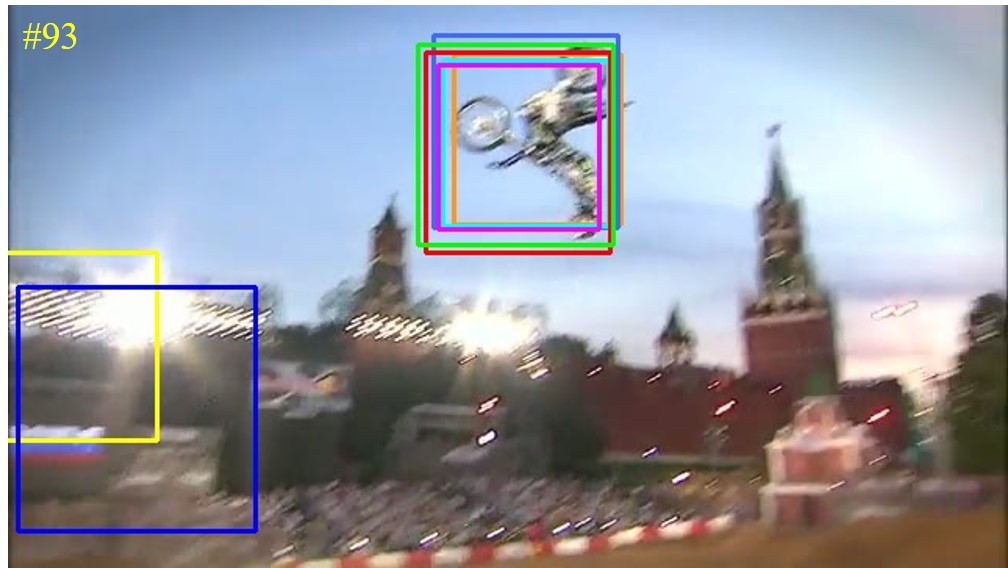}&
		\includegraphics[width=\wdenoising, height=\hdenoising]{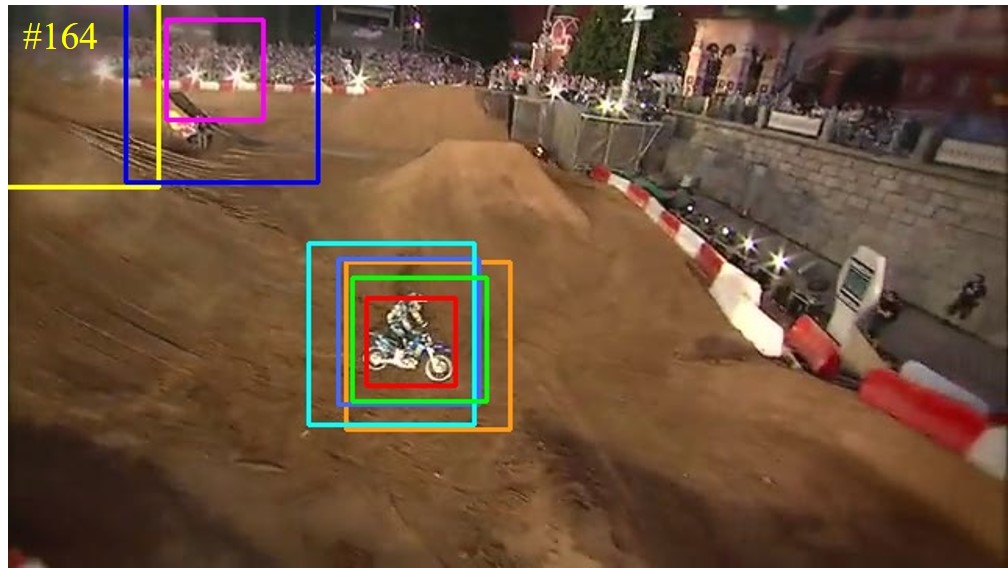}\\
		
		\includegraphics[width=\wdenoising, height=\hdenoising]{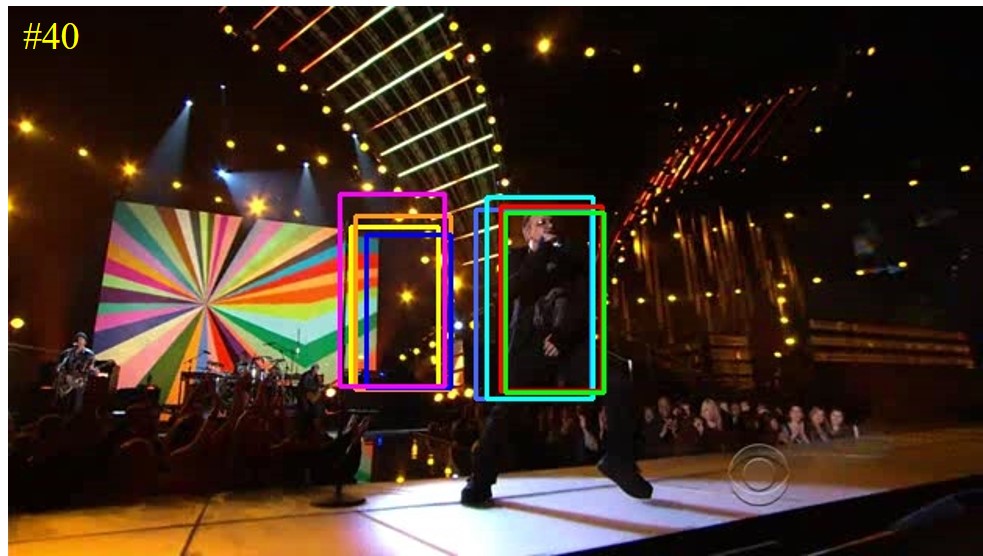}&
		\includegraphics[width=\wdenoising, height=\hdenoising]{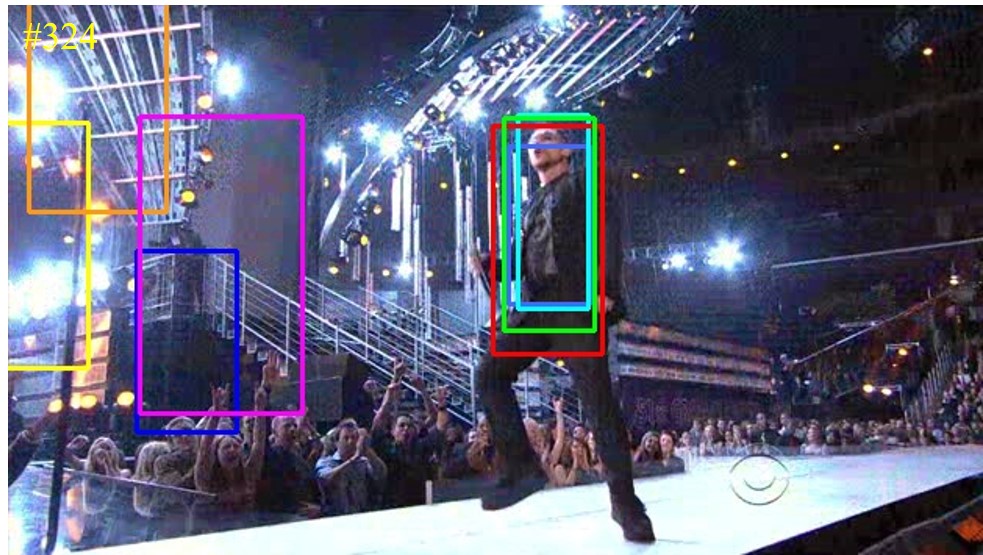}&
		\qquad   &
		\includegraphics[width=\wdenoising, height=\hdenoising]{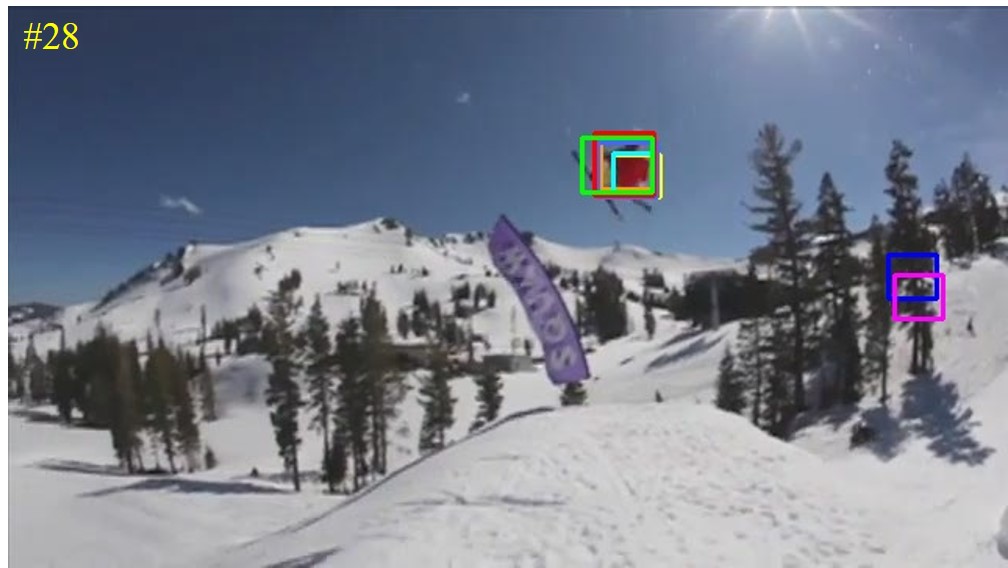}&
		\includegraphics[width=\wdenoising, height=\hdenoising]{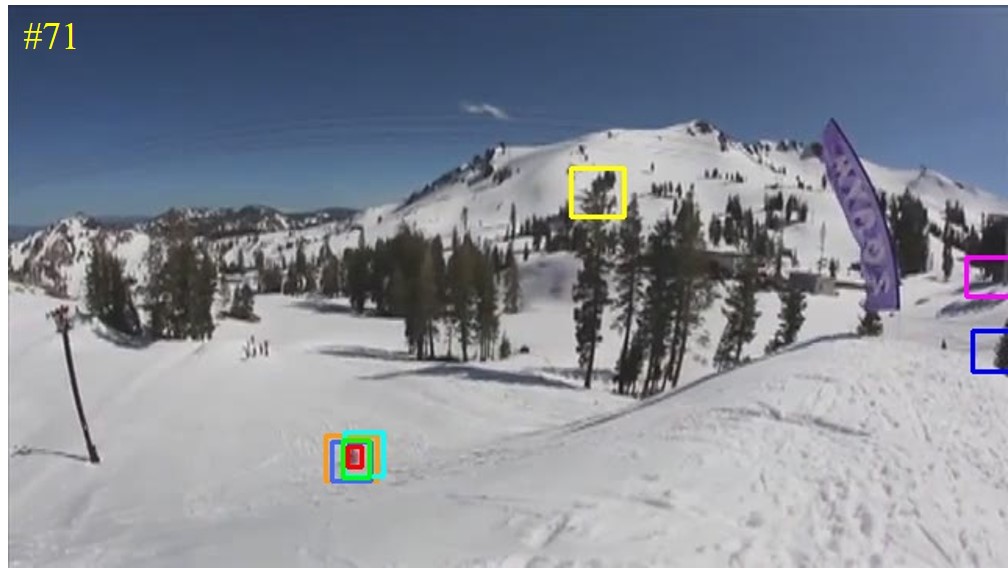}\\
		
		\includegraphics[width=\wdenoising, height=\hdenoising]{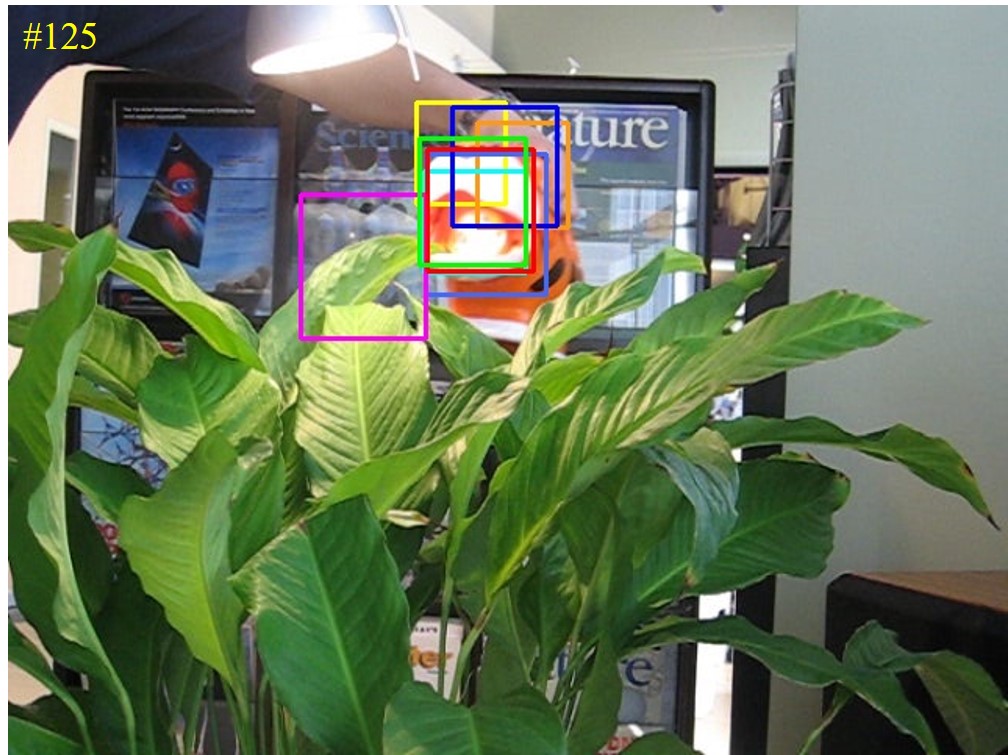}&
		\includegraphics[width=\wdenoising, height=\hdenoising]{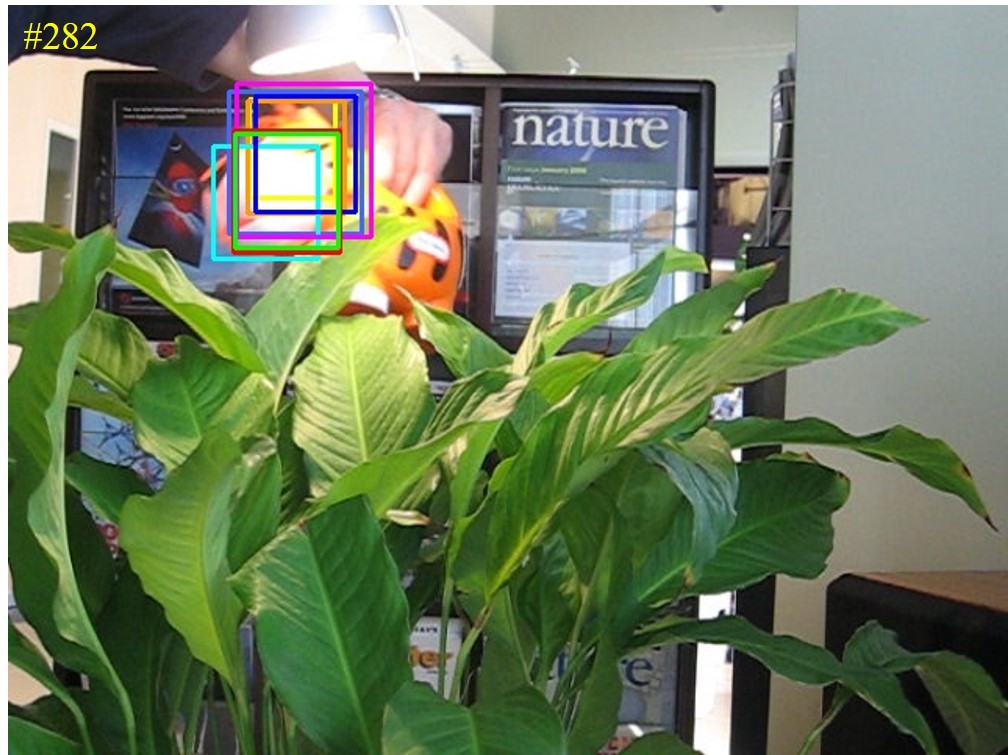}&
		\qquad   &
		\includegraphics[width=\wdenoising, height=\hdenoising]{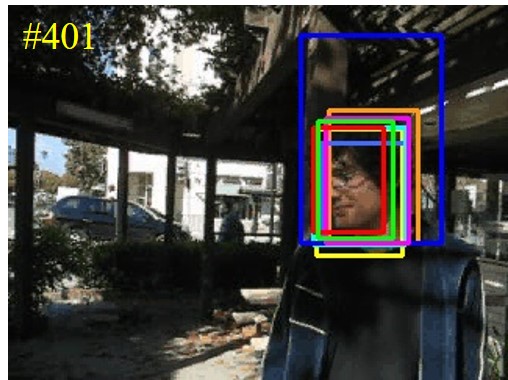}&
		\includegraphics[width=\wdenoising, height=\hdenoising]{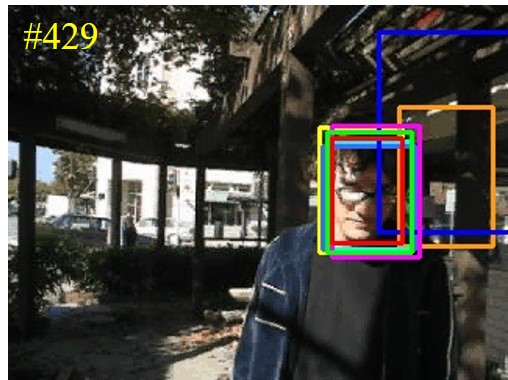}\\
		
		\multicolumn{5}{c}{\includegraphics[width=\wcolor, height=\hcolor]{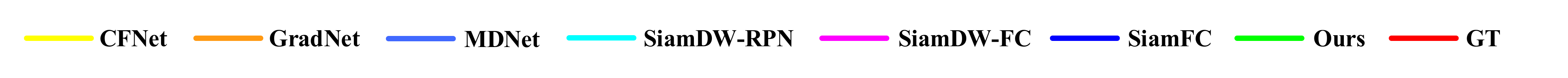}} \\
		\vspace{-0.8cm}
	\end{tabular}
	\caption{Qualitative evaluation of our  method and other
		trackers including CFNet~\cite{valmadre2017end},GradNet~\cite{li2019gradnet}, MDNet~\cite{nam2016learning},   SiamDW-RPN~\cite{zhang2019deeper}, SiamDW-FC~\cite{zhang2019deeper}, and SiamFC~\cite{bertinetto2016fully} on 8 challenging videos from OTB2013~\cite{wu2015object}. From left to right and	top to down are \textit{Ironman}, \textit{CarScale}, \textit{Matrix}, \textit{MotorRolling}, \textit{Skating1}, \textit{Skiing}, \textit{Tiger2}, and \textit{Trellis}  respectively. Best viewed in zoom in.}
	\label{fig:example_event}
\end{figure*}

\subsection{\zjq{Evaluation on Standard RGB Benchmark}}
To demonstrate the effectiveness of our MCFR, we first test it on the standard RGB benchmark OTB2013~\cite{wu2015object}.
The evaluation is based on two metrics: the Precision Rate (PR) and the Success Rate (SR). SR cares the frame of that overlap between ground truth and predicted bounding box is larger than a threshold; PR focuses on the frame of that the center distance between ground truth and predicted bounding box within a given threshold. The one-pass evaluation (OPE) is employed to compare our algorithm with the eleven state-of-the-art trackers including SiamDW-RPN~\cite{zhang2019deeper}, MDNet~\cite{nam2016learning}, SiamFC~\cite{bertinetto2016fully}, CFNet~\cite{valmadre2017end}, SiamRPN~\cite{li2018high},   SiamDW-FC~\cite{zhang2019deeper}, DaSiamRPN~\cite{zhu2018distractor}, SRDCF~\cite{danelljan2015learning}, GradNet~\cite{li2019gradnet}, DiMP~\cite{bhat2019learning}, and ATOM~\cite{danelljan2019atom}. We also apply ESIM~\cite{rebecq2018esim} to generate event-based data on OTB2013~\cite{wu2015object}.

The evaluation results are reported in Figure~\ref{fig:prsr}. From the results, we can see that our method outperforms the other trackers on OTB2013~\cite{wu2015object}. In particular, our MCFR (95.3\%/73.9\% in PR/SR) outperforms 3.1\% over the second-best tracker MDNet~\cite{nam2016learning} in SR, and is superior to other trackers in PR. It demonstrates the effectiveness of our structure for extracting the common features and unique features from different domains. In addition, the remarkable superior performance over the state-of-the-art trackers like ATOM~\cite{danelljan2019atom} and DiMP~\cite{bhat2019learning} suggests that our method is able to make the best use of event domain information to boost tracking performance.

In order to analyze what reliable information the event-based data provides, we report the results on various challenge attributes to show more detailed performance. As shown in Figure~\ref{fig:OTB2013_various}, our tracker can effectively handle these challenging situations that traditional RGB trackers often lose targets. In particular, under the challenging scenes of fast motion and motion blur, our tracker greatly surpasses the other trackers. That's because the low latency and high temporal resolution of the event-based camera bring more information about the movement between adjacent RGB frames, which can effectively promote the performance of our tracker. From Figure~\ref{fig:OTB2013_various}, we can also find that our tracker has the best performance
in illumination variation scenes. Moreover, in the \textit{background\_clutter} (the background near the target has the similar color or texture as the target), as event-based data pays more attention to moving objects rather than the color or texture of objects, our tracker has been significantly improved.

\setlength{\tabcolsep}{13.0pt}
\begin{table*}[t]
	\caption{Results obtained by the competitors and our method on the EED~\cite{mitrokhin2018event} dataset. The best results are in \textcolor{red}{red}.}
	\label{tab:2}
	\begin{tabular}{c|cc|cc|cc|cc}
		
		\hline   
		\hline
		\multirow{2}{*}{Methods} &  \multicolumn{2}{|c|}{fast\_drone} &  \multicolumn{2}{|c|}{light\ variations}  & \multicolumn{2}{|c|}{what is background} & \multicolumn{2}{|c}{occlusions}  \\
		\cline{2-9} 
		
		&AP$~\uparrow$ &AR$~\uparrow$  &AP$~\uparrow$ &AR$~\uparrow$ &AP$~\uparrow$ &AR$~\uparrow$ &AP$~\uparrow$ &AR$~\uparrow$ \\
		
		\hline
		\hline                  
		KCL~\cite{henriques2014high} & 0.169 &0.176 & 0.107&0.066 &0.028 &0.000 &0.004 &0.000 \\
		TLD~\cite{kalal2011tracking} & 0.315 &  0.118& 0.045& 0.066& 0.269& 0.333& 0.092& 0.167\\
		SiamFC~\cite{bertinetto2016fully} &0.559 & 0.667 & 0.599&  0.675&0.307 &0.308 &0.148 &0.000 \\
		ECO~\cite{danelljan2017eco} &0.637 &0.833 &0.586 &0.688  &0.616 &0.692 & 0.108 &0.143 \\
		DaSiamRPN~\cite{zhu2018distractor} &0.673 & 0.853&0.654  &0.894  &0.678 &0.833 &0.189  &0.333 \\
		E-MS~\cite{barranco2018real} &0.313  &0.307 & 0.325 & 0.321 & 0.362&0.360 &0.356 &0.353 \\
		ETD~\cite{chen2019asynchronous} &0.738 & 0.897 &0.842  &\textcolor{red}{0.933 }& 0.653 &0.807 & 0.431&\textcolor{red}{0.647} \\
		MCFR(Ours) &\textcolor{red}{0.802} &\textcolor{red}{0.931} &\textcolor{red}{0.853} &\textcolor{red}{0.933} &\textcolor{red}{0.734} &\textcolor{red}{0.871} &\textcolor{red}{0.437} &0.644 \\		
		\hline
		\hline
	\end{tabular}
\end{table*}
\subsection{\zjq{Evaluation on Real Event Dataset}}
To further prove the effectiveness of our method, we also evaluate it on the real event dataset EED~\cite{mitrokhin2018event}. The EED~\cite{mitrokhin2018event} was recorded using a DAVIS~\cite{brandli2014a} event camera in real-world environments, which contains the events sequences and the corresponding RGB sequences for each video. The EED~\cite{mitrokhin2018event} also provides the ground truth for targets. The EED~\cite{mitrokhin2018event} contains five sequences: \textit{fast\_drone}, \textit{light\_variations}, \textit{what\_is\_background}, \textit{occlusions}, and \textit{multiple\_objects}. Since \textit{multiple\_objects} involves multiple targets, we use the first four video sequences here. Specifically, \textit{fast\_drone} describes a fast moving drone under a very low illumination condition, and in \textit{light\_variations}, a strobe light flashing at a stable frequency is placed in a dark room. A thrown ball with a dense net as foreground in \textit{what\_is\_background}, and a thrown ball with a short occlusion under a dark environment in \textit{occlusions}.

Following~\cite{chen2019asynchronous}, we use two metrics: the Average Precision (AP) and the Average Robustness (AR) for evaluation. 
AP and AR describe the accuracy and robustness of the tracker, respectively.
The AP can be formulated as follows:
\begin{equation}
AP = \frac{1}{N}\frac{1}{M}\sum_{a=1}^{N}\sum_{b=1}^{M}\frac{O_{a,b}^{E}\cap O_{a,b}^{G}}{O_{a,b}^{E} \cup  O_{a,b}^{G}} ,
\end{equation}
where $N$ is the repeat times of the evaluation (here we set $N$ to 5), and $M$ is the number of objects in the current sequence. $O_{a,b}^{E}$ is the estimated bounding box in the $a$-th round of the evaluation for the $b$-th object, and  $O_{a,b}^{G}$ is the corresponding ground truth. The AR can be formulated as follows:
\begin{equation}
AR = \frac{1}{N}\frac{1}{M}\sum_{a=1}^{N}\sum_{b=1}^{M}success_{a,b} ,
\end{equation}
where $success_{a,b}$ indicates that whether the tracking in the $a$-th round for the $b$-th object is successful or not. It will be considered a failure condition if the AP value is less than 0.5.
We compare our algorithm with seven state-of-the-art methods including 
KCL~\cite{henriques2014high}, TLD~\cite{kalal2011tracking}, SiamFC~\cite{bertinetto2016fully}, ECO~\cite{danelljan2017eco}, DaSiamRPN~\cite{zhu2018distractor}, E-MS~\cite{barranco2018real}, and ETD~\cite{chen2019asynchronous}. Herein, the first five algorithms are correlation filter-based or deep learning based traditional RGB object tracking methods, and the remaining are event-based tracking methods.
\zjq{The quantitative results are shown in Table~\ref{tab:2}, we can see that the traditional RGB trackers are severely affected by low light and fast motion.} When there is too much noise in the events, due to lacking image texture information, the event-based tracker cannot effectively obtain satisfactory performance. Instead, our proposed structure can simultaneously obtain texture information from RGB and target edge cues from events so that our method can effectively handle high dynamic range and fast motion conditions.

%
%
%
%

\setlength{\tabcolsep}{1.6pt}
\begin{table}[t]
	\small
	\centering

	\caption{Ablation analyses of MCFR and its variants.}
	\label{tab:3}
	\begin{tabular}{l|ccc|cc|cc|cc}
		\hline     
		\hline   
		& UEE & CFE & UER &RGB &Event&C&T & PR(\%) & SR(\%) \\
		\hline
		MCFR$_{oE}$ & $\times$  &  \checkmark & $\times$ &$\times$  &\checkmark  &  \checkmark  &  \checkmark &0.397 &0.556  \\
		MCFR$_{oR}$ & $\times$  &  \checkmark & $\times$ &\checkmark  &$\times$  &   $\times$  &  $\times$ &0.702 &0.944  \\
		MCFR$_{ER}$ & $\times$  &  \checkmark & $\times$ &\checkmark  &\checkmark  &  \checkmark  &  \checkmark &0.719 &0.949 \\
		\hline
		
		w/o \textit{UEE} & $\times$  & \checkmark  & \checkmark & \checkmark  &\checkmark  &  \checkmark  &  \checkmark &0.729 &0.950 \\
		w/o \textit{CFE} & \checkmark  &  $\times$ &  \checkmark & \checkmark& \checkmark &  \checkmark  &  \checkmark &0.710  &0.947 \\
		w/o \textit{UER} & \checkmark  &  \checkmark & $\times$ &\checkmark  &\checkmark  &  \checkmark  &  \checkmark
		&0.723 &0.951  \\
		\hline
		MCFR$_{C}$ & \checkmark  &  \checkmark &  \checkmark &\checkmark  &\checkmark  &  \checkmark  &  $\times$ &0.720 &0.951  \\
		MCFR$_{T}$ &  \checkmark &  \checkmark &  \checkmark &\checkmark  &\checkmark  &  $\times$  &  \checkmark &0.728 &0.950  \\
		\hline
		MCFR & \checkmark  & \checkmark  & \checkmark  &\checkmark &\checkmark  &  \checkmark  &  \checkmark& \textcolor{red}{0.739} &\textcolor{red}{0.953} \\
		\hline
		\hline
	\end{tabular}
\end{table}

\subsection{\zjq{Ablation Study}}
To verify RGB images and event-based data can jointly promote the tracker performance, we implement three variants, including 1) MCFR$_{oE}$, that only applies CFE with events as inputs. 2) MCFR$_{oR}$, that only applies CFE with RGB images as inputs.
3) MCFR$_{ER}$, that applies CFE with events and RGB data as inputs. The comparison results are shown in Table~\ref{tab:3}. The results illustrate that the collaborative use of multi-domain information is indeed superior to a single domain.

To validate our method can effectively extract common and unique features from RGB and event domains, we implement three variants based on MCFR, including 1) w/o \textit{UEE}, that removes Unique Extractor for Event, 2) w/o \textit{CFE}, that removes Common Feature Extractor, and 3) w/o \textit{UER}, that removes Unique Extractor for RGB.
From Table~\ref{tab:3}, we can see that our MCFR is superior over  w/o \textit{UEE}, which suggests the UEE with SNNs is helpful to take advantage of the event-based data, thereby improving the tracking performance. Besides, MCFR outperforms w/o \textit{CFE} by a clear margin demonstrates that it is essential to extract common features of targets. The superior performance of MCFR over w/o \textit{UER} suggests unique texture features from RGB are important for tracking.

We also explore the performance impact of different ways of stacking events.  MCFR$_{C}$ and  MCFR$_{T}$ represent stacking event streams according to counts and the latest timestamp, respectively. From Table~\ref{tab:3}, we can see that MCFR outperforms MCFR$_{C}$ and  MCFR$_{T}$, which verifies that counts images \textit{C} can record all the events that occurred within a period, and timestamps images \textit{T} can encode features about the motion.

\subsection{\zjq{Failure Cases Analysis}}

Our method does have limitations. 
The failure examples are shown in Figure~\ref{fig:failure_case}. Since the target is static, the event camera cannot effectively provide the edge cues of the target, resulting in the unavailability of information in the event domain. At the same time, an object similar to the target moves around the target, similar colors and textures will interfere with the target-related information provided by the RGB domain. In these cases, the event provides misleading information about moving object, which causes incorrect positioning.
\def\wdenoising{0.5\linewidth}
\def\hdenoising{1.1in}
\begin{figure}[t]
	\setlength{\tabcolsep}{2.4pt}
	\centering
	\begin{tabular}{cc}
		
		\includegraphics[width=\wdenoising, height=\hdenoising]{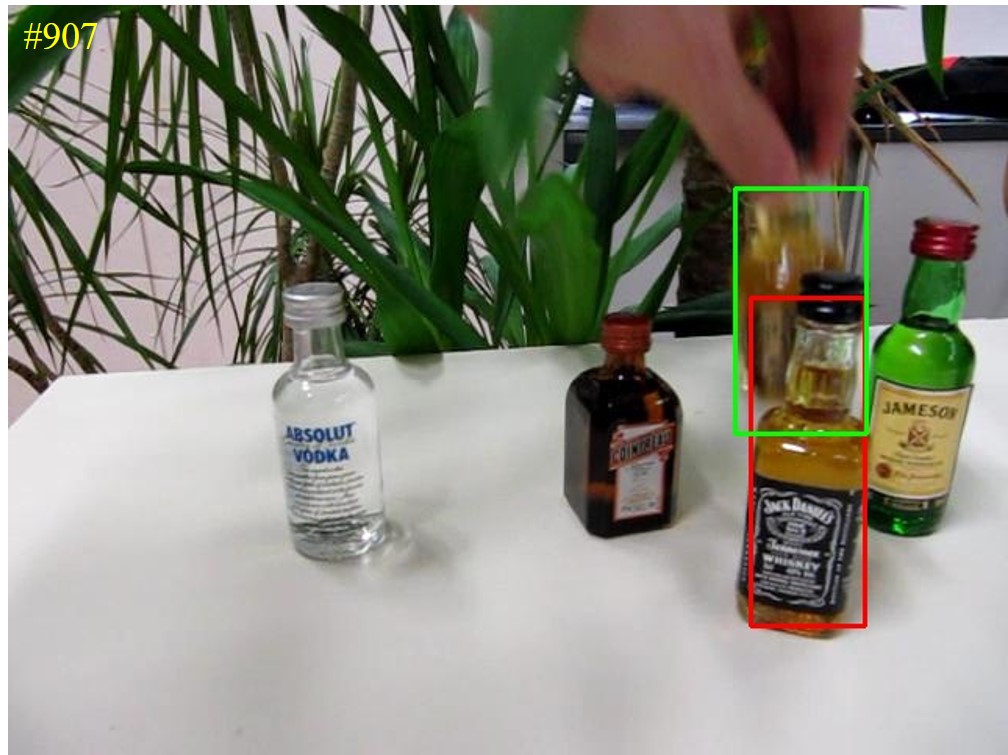}&
		\includegraphics[width=\wdenoising, height=\hdenoising]{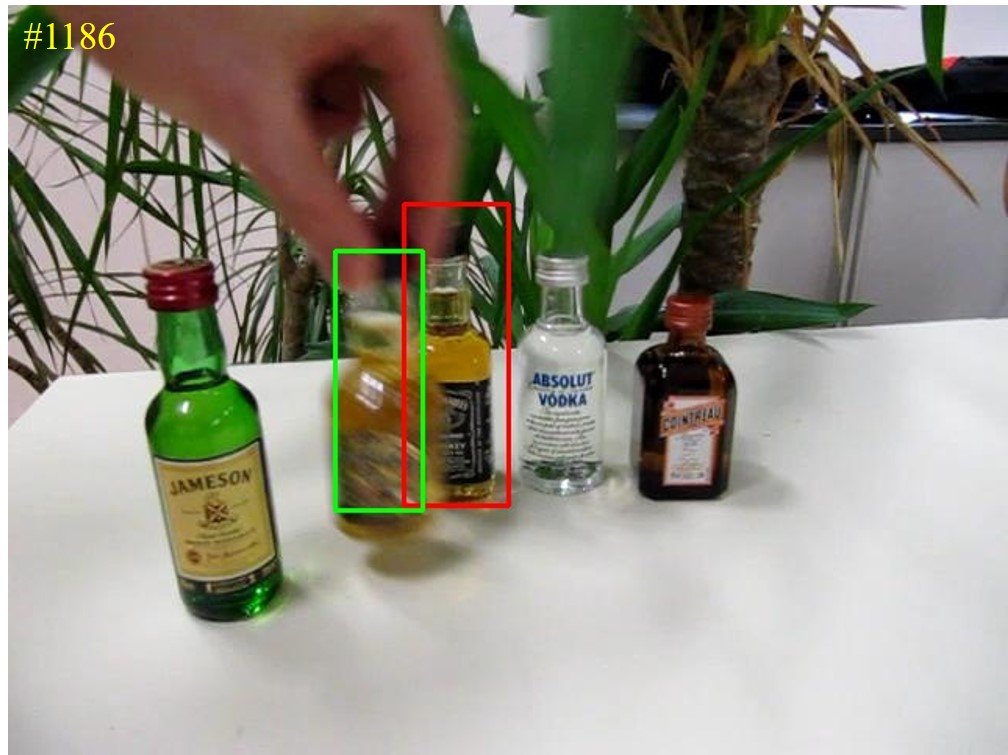} \\
	\end{tabular}
	\caption{Failure cases. The target is stationary and a moving object similar to the target appears around. Red box is GT, green box is our result.}
	\label{fig:failure_case}
\end{figure}

\section{\zjq{Conclusion}}

In this paper, we propose Multi-domain Collaborative Feature Representation (MCFR) to effectively extract and fuse common features and unique features from the RGB and event domain for robust visual object tracking in some challenging conditions, such as fast motion and high dynamic range. 
Specifically, we apply CFE to extract common features and design UEE based on SNNs and UER based on DCNNs to present specific features of the RGB and event data. 
Extensive experiments on the RGB tracking benchmark and real event dataset  suggest that the proposed tracker achieves outstanding performance. 
In future work, we will explore upgrading our event-based module so that it can be easily extended to existing RGB trackers for improving performance in challenging conditions.

\noindent
\textbf{Acknowledgements.} This work was supported in part by the National Natural Science Foundation of China under Grant 91748104, Grant 61972067, and the Innovation Technology Funding of Dalian (Project No. 2018J11CY010, 2020JJ26GX036).

\noindent
\textbf{Conflict of interest.} Jiqing Zhang, Kai Zhao, Bo Dong, Yingkai Fu, Yuxin Wang, Xin Yang and Baocai Yin declare that
they have no conflict of interest.
\bibliographystyle{spmpsci}
\bibliography{bib_CGIconf} 

\end{document}